\def\year{2021}\relax
\documentclass[letterpaper]{article} 
\usepackage{aaai21}  
\usepackage{times}  
\usepackage{helvet} 
\usepackage{courier}  
\usepackage[hyphens]{url}  
\usepackage{graphicx} 
\usepackage{natbib}  
\usepackage{caption} 
\usepackage{colortbl}
\newcommand{\CC}[1]{\cellcolor{blue!#1}}
\usepackage[table, svgnames, dvipsnames]{xcolor}

\usepackage{latexsym}

\usepackage{booktabs} 
\usepackage{multirow}
\usepackage{microtype}
\usepackage{amsfonts}
\usepackage{amsmath} 

\usepackage{caption}
\usepackage{subcaption}
\usepackage{graphicx}

\usepackage[raggedright]{sidecap} 

\usepackage{arydshln}

\usepackage{comment}
\usepackage{float}
\usepackage[ruled,vlined,linesnumbered]{algorithm2e}
\usepackage{algpseudocode}
\usepackage{xspace}

\usepackage{CJKutf8}
\usepackage{float}

\usepackage{cleveref}
\crefformat{section}{\S#2#1#3}
\crefformat{subsection}{\S#2#1#3}
\crefformat{subsubsection}{\S#2#1#3}
\crefrangeformat{section}{\S\S#3#1#4 to~#5#2#6}
\crefmultiformat{section}{\S\S#2#1#3}{ and~#2#1#3}{, #2#1#3}{ and~#2#1#3}
\usepackage{refstyle}
\Crefformat{figure}{#2Fig.~#1#3}
\Crefmultiformat{figure}{Figs.~#2#1#3}{ and~#2#1#3}{, #2#1#3}{ and~#2#1#3}
\Crefformat{table}{#2Tab.~#1#3}
\Crefmultiformat{table}{Tabs.~#2#1#3}{ and~#2#1#3}{, #2#1#3}{ and~#2#1#3}

\newcommand{\modelname}[0]{\textbf{\textsc{EVA}}\xspace}

\newcommand{\stitle}[1]{\vspace{0.3ex} \noindent{\bf #1}}

\title{Visual Pivoting for (Unsupervised) Entity Alignment}

\author {
    Fangyu Liu\textsuperscript{\rm 1},
    Muhao Chen\textsuperscript{\rm 2,3},
    Dan Roth\textsuperscript{\rm 2}, 
    Nigel Collier\textsuperscript{\rm 1}  \\
}
\affiliations {
    \textsuperscript{\rm 1} Language Technology Lab, TAL, University of Cambridge, UK \\
    \textsuperscript{\rm 2} Department of Computer and Information Science, University of Pennsylvania, USA \\
    \textsuperscript{\rm 3} Viterbi School of Engineering, University of Southern California, USA \\
    \texttt{\{fl399, nhc30\}@cam.ac.uk}, 
    \texttt{muhaoche@usc.edu}, 
    \texttt{danroth@seas.upenn.edu}
}
\begin{document}

\maketitle

\begin{abstract}
This work studies the use of visual semantic representations to align entities in heterogeneous knowledge graphs 
(KGs). Images are natural components of many existing KGs. By combining visual knowledge with other auxiliary information,
we show that 
the proposed new approach,  \modelname
\includegraphics[height=1em]{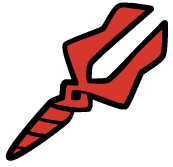}
, creates a holistic entity representation that provides strong signals for cross-graph entity alignment. 
Besides, 
previous entity alignment methods require human labelled seed alignment, restricting availability.
\modelname provides a completely unsupervised solution by leveraging the visual similarity of entities to create an initial seed dictionary (visual pivots). 
Experiments on benchmark data sets DBP15k and DWY15k show that \modelname offers state-of-the-art performance on both monolingual and cross-lingual entity alignment tasks.
Furthermore, we discover 
that images are particularly useful to align long-tail KG entities, which inherently lack the structural contexts that are necessary for capturing the correspondences.\footnote{Code release: \url{https://github.com/cambridgeltl/eva}. Project page: \url{http://cogcomp.org/page/publication_view/927}.}
\end{abstract}

\section{Introduction}


Knowledge graphs (KGs) such as DBpedia \citep{lehmann2015dbpedia}, YAGO \citep{rebele2016yago} and Freebase \citep{bollacker2008freebase} 
store structured knowledge that is crucial to
numerous knowledge-driven applications including question answering \citep{cui2017kbqa}, entity linking \citep{radhakrishnan2018elden}, text generation \citep{koncel2019text} and information extraction \citep{hoffmann2011knowledge}.  
However, 
most KGs are independently extracted from separate sources, 
or contributed by speakers of one language, therefore limiting the coverage of knowledge. 
It is important to match and synchronise the independently built KGs and seek to provide NLP systems the benefit of complementary information contained in different KGs \citep{bleiholder2009data,bryl2014learning}.
To remedy this problem, the Entity Alignment (EA)\footnote{The \emph{entity} in EA refers to real-world objects and concepts.} task aims at building cross-graph mappings 
to match 
entities 
having the same real-world identities,
therefore integrating knowledge from different sources into a common space.

\begin{figure*}
    \centering
    \small
    \includegraphics[width=0.8\textwidth]{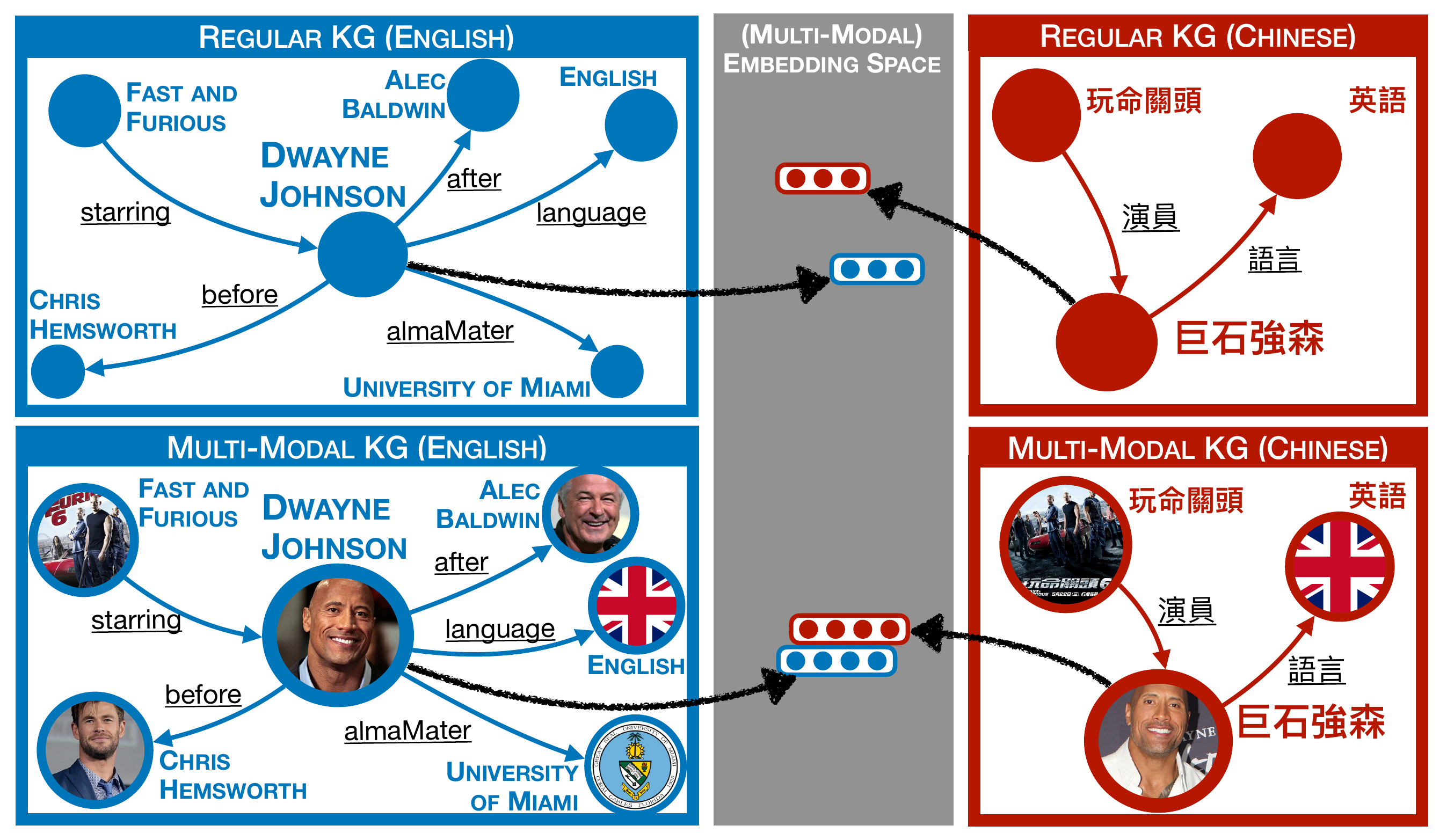}
    \caption{\small{A challenging example which heavily benefits from using vision to align entities between cross-lingual KGs in DBP15k. We display the neighbourhoods of entity \textsc{Dwayne\_Johnson} (\begin{CJK*}{UTF8}{bsmi}巨石強森\end{CJK*}) in English and Chinese KGs. Without the visual component, the two embedded vectors are far apart (top) while similar images can pull them together (bottom).}}
    \label{fig:intro}
\end{figure*}


A major bottleneck
for training EA models is the scarce
cross-graph pivots\footnote{In this paper, \emph{pivot} is used interchangeably with \emph{seed alignment} between cross-graph entities; \emph{visual pivoting} means to use the visual space as intermediate to find seed alignment.} 
available as alignment signals
\cite{chen2017multigraph,sun2018bootstrapping}. Besides, the sparsity of KGs 
is usually accompanied with weak structural correspondence,
posing an even greater challenge to EA. 
To mitigate this problem,
recent works have attempted to retrieve auxiliary supervision signals from the supplementary 
information of entities, such as attributes \citep{sun2017cross,trsedya2019attr,yang2020cotsae,liu2020exploring} and descriptions \citep{chen2018co}. 
However, existing EA approaches are still limited in their capabilities. Our study proposes to leverage images, a natural component of entity profiles in many KGs \cite{lehmann2015dbpedia,vrandevcic2014wikidata,liu2019mmkg}, for better EA. Images have been used to enrich entity representations for KG completion in a single-graph scenario \citep{xie2017image,moussellySergieh2018multimodal,pezeshkpour2018embedding}. However, the visual modality is yet to be explored for cross-graph tasks such as EA.


Our study stands upon several advantages that the visual modality brings to EA.
First, the visual concept of a named entity is usually universal, regardless of the language or the schema of the KG. Therefore,
given a well-designed algorithm, images should provide the basis to find a set of reliable pivots.
Second, images in KGs are freely-available and of high quality. Crucially, they are mostly manually verified and disambiguated. These abundant gold visual attributes in KGs render EA an ideal application scenario for visual representations. 
Third, images offer the possibility to enhance the representation of rare KG entities with impoverished structural contexts \citep{cao2020open,xiong2018one,hao2019joie}.
Images can be particularly beneficial in this setting, as entities of lower frequencies tend to be more concrete concepts \citep{hessel2018quantifying} with stable visual representations \citep{kiela2014improving,hewitt2018learning}. 
To demonstrate the benefit from injecting images,
we present a challenging example in  \Cref{fig:intro}. Without images, it is harder to infer the correspondence between \textsc{Dwayne\_Johnson} and its counterpart \begin{CJK*}{UTF8}{bsmi}巨石強森\end{CJK*} (``The Rock Johnson'') due to their dissimilar
neighbourhoods in the two KGs.
An alignment can be more easily induced by detecting visual similarity.

In this work, 
we propose \modelname (\textbf{\textsc{E}}ntity \textbf{\textsc{V}}isual \textbf{\textsc{A}}lignment), which incorporates 
images along with structures, relations and attributes to align entities in different KGs. During training, a learnable attention weighting scheme helps the alignment model to decide on the importance of each modality, and also provides interpretation for each modality's contribution.
As we show, an advantage of our approach is that the model is able to be trained on either a small set of seed alignment labels as in previous methods 
(semi-supervised setting), or using only a set of automatically induced visual pivots (unsupervised setting). Iterative learning (\textsc{il}) is applied to expand the set of training pivots under both settings. 
On two large-scale standard benchmarks, i.e. DBP15k 
for cross-lingual EA and DWY15k 
for monolingual EA,
\modelname variants with or without alignment labels consistently outperform competitive baseline approaches. 

The contributions of this work are three-fold: (i) We conduct the first investigation into the use of images as part of entity representations for EA, and achieve state-of-the-art (SOTA) performance across all settings. (ii) We leverage visual similarities to propose a fully unsupervised EA setting, avoiding reliance on any gold labels. Our model under the unsupervised setting performs closely to its semi-supervised results, even surpassing the previous best semi-supervised methods. 
(iii) We offer interpretability in our study by conducting ablation studies on the contributions from each modality and a thorough error analysis. We also provide insights on images' particular impact on long-tail KG entities. 

\section{Related Work}
Our work is connected to two 
research topics.
Each has a large body of work which we can only provide as a highly selected summary.

\stitle{Entity alignment.}
While early work employed symbolic or schematic methods to address the EA problem \citep{wijaya2013pidgin,suchanek2011paris},
more recent attention has been paid to embedding-based methods.
A typical method of such is \textsc{MTransE} \citep{chen2017multigraph}, which jointly trains a translational embedding model \citep{bordes2013translating} to encode language-specific KGs in separate embedding spaces, and a transformation to align counterpart entities across embeddings. 
Following this methodology, later works span the following three lines of studies to improve on this task.
The first is to use alternatives of embedding learning techniques.
Those include more advanced relational models such as contextual translations \citep{sun2019transedge}, residual recurrent networks (\textsc{Rsn}, \citealt{guo2019learning}) and relational reflection transformation \citep{mao2020relational}, as well as variants of graph neural networks (GNNs) such as \textsc{Gcn} \citep{wang2018cross,yang2019aligning,wu2019relation,wu2019jointly}, \textsc{Gat} \citep{sun2020alinet,zhu2019neighborhood} and multi-channel GNNs \citep{cao2019multi}.
The second line of research focuses on capturing the alignment of entities with limited labels, therefore incorporating semi-supervised or metric learning techniques such as bootstrapping \citep{sun2018bootstrapping}, co-training \citep{chen2018co,yang2020cotsae} and optimal transport \citep{pei2019transport}.
Besides, to compensate for limited supervision signals in alignment learning, another line of recent works retrieves auxiliary supervision from side information of entities.
Such information include numerical attributes \citep{sun2017cross,trsedya2019attr}, literals \citep{zhang2019multi,otani2018cross} and descriptions of entities \citep{chen2018co,gesese2019survey}.
A recent survey by \citet{sun2020benchmark} has systematically summarised works in these lines.

The main contribution of this paper is relevant to the last line of research.
To the best of our knowledge, 
this is the first attempt to incorporate the visual modality for EA in KGs.
It also presents an effective unsupervised solution to this task, without the need of alignment labels that are typically required in previous works.

\stitle{Multi-modal KG embeddings.} While incorporating perceptual qualities has been a hot topic for language representation learning for many years, few attempts have been made towards building multi-modal KG embeddings. \citet{xie2017image} and \citet{thoma2017towards} are among the first to 
incorporate translational KG embedding methods \citep{bordes2013translating} with external visual information. However, they mostly explore the joint embeddings on intrinsic tasks like word similarity and link prediction. \citet{moussellySergieh2018multimodal} improve the model of \citeauthor{xie2017image} to incorporate both visual and linguistic features under a unified translational embedding framework. \citet{pezeshkpour2018embedding} and \citet{rubio2019answering} also model the interplay of images and KGs. However, \citeauthor{pezeshkpour2018embedding} focus specifically on KG completion. \citeauthor{rubio2019answering} treat images as first class citizens for tasks like answering vision-relational queries instead of building joint representation for images and entities. 
The aforementioned works all focus on single KG scenarios.
As far as we know,
we are the first to use the intermediate visual space for EA between KGs.


Note that in the context of embedding alignment, many studies have incorporated images in lexical or sentential representations to solve cross-lingual tasks such as bilingual lexicon induction \citep{vulic2016multi,rotman2018bridging,sigurdsson2020visual} or cross-modal matching tasks such as text-image retrieval \citep{gella2017image,kiros2018illustrative,kiela2018dynamic}. Beyond embedding alignment, the idea of visual pivoting is also popular in the downstream task of machine translation \citep{caglayan2016does,huang2016attention,hitschler2016multimodal,specia2016shared,calixto2017incorporating,barrault2018findings,su2019unsupervised}, but it is beyond the scope of this study. All of these works are not designed to deal with relational data 
that are crucial to performing EA. 

\section{Method}
\label{sec:method}

We start describing our method by formulating the learning resources. 
A KG ($\mathcal{G}$) 
can be viewed as a set of triplets that are constructed with an entity vocabulary ($\mathcal{E}$) and a relation vocabulary ($\mathcal{R}$),
i.e. 
$\mathcal{G} = \{(e_1,r,e_2):r \in \mathcal{R}; e_1,e_2\in \mathcal{E}\}$ where a triplet records the relation $r$ between the head and tail entities $e_1,e_2$. 
Let $\mathcal{G}_s = \mathcal{E}_s\times \mathcal{R}_s \times \mathcal{E}_s,\mathcal{G}_t = \mathcal{E}_t\times \mathcal{R}_t \times \mathcal{E}_t$ denote two individual KGs (to be aligned). Given a pair of entities $e_s\in\mathcal{E}_s$ from source KG and $e_t\in\mathcal{E}_t$ from target KG, the goal of EA is to learn a function $f(\cdot,\cdot;\theta): \mathcal{E}_s \times \mathcal{E}_t \rightarrow \mathbb{R}$ parameterised by $\theta$ that can estimate the similarity of $e_s$ and $e_t$. $f(e_s,e_t; \theta)$ should be high if $e_s,e_t$ are describing the same identity and low if they are not. 
Note that $\mathcal{E}_s$ and $\mathcal{E}_t$ ensure 1-to-1 alignment \cite{chen2018co,sun2018bootstrapping}, as to be congruent to the design of mainstream KBs \cite{lehmann2015dbpedia,rebele2016yago} where disambiguation of entities is granted. To build joint representation for entities, we consider auxiliary information including images, relations and attributes. Let $\mathcal{I}$ denote the set of all images; $\mathbf{R}\in\mathbb{R}^{N\times d_R},\mathbf{A}\in\mathbb{R}^{N\times d_A}$ denote the matrices of relation and attribute features. 

To tackle the EA task, our method jointly conducts two learning processes.
A multi-modal embedding learning process aims at encoding both KGs $\mathcal{G}_s$ and $\mathcal{G}_t$ in a shared embedding space.
Each entity in the embedding space is characterised based on both the KG structures and auxiliary information including images.
In the shared space, the alignment learning process seeks to precisely capture the correspondence between counterpart entities by 
Neighbourhood Component Analysis (NCA, \citealt{goldberger2005neighbourhood,liu2020hal}) and iterative learning.
Crucially, the alignment learning process can be unsupervised, i.e. pivots are automatically inferred from the visual representations of entities without the need of EA labels.
The rest of this section introduces the technical details of both learning processes.

\subsection{Multi-modal KG Embeddings}
\label{sec:method_MMKGE}

Given entities from two KGs $\mathcal{G}_s$ and $\mathcal{G}_t$, and the auxiliary data $\mathcal{I},\mathbf{R},\mathbf{A}$, this section details how they are embedded into low-dimensional vectors. 

\stitle{Graph structure embedding.}
To model the structural similarity of $\mathcal{G}_s$ and $\mathcal{G}_t$, capturing both entity and relation proximity, we use Graph Convolutional Network (\textsc{Gcn}) proposed by \citet{kipf2017semi}. Formally, a multi-layer \textsc{Gcn}'s operation on the $l$-th layer can be formulated as: 
\begin{equation}
    \mathbf{H}^{(l+1)} = [ \ \Tilde{\mathbf{D}}^{-\frac{1}{2}}\Tilde{\mathbf{M}}\Tilde{\mathbf{D}}^{-\frac{1}{2}}\mathbf{H}^{(l)}\mathbf{W}^{(l)} \ ]_+,
\end{equation}
where $[\cdot]_+$ is the ReLU activation; $\Tilde{\mathbf{M}}=\mathbf{M}+\mathbf{I}_N$ is the adjacency matrix of
$\mathcal{G}_s \cup \mathcal{G}_2$ plus an identity matrix (self-connection); $\Tilde{\mathbf{D}}$ is a trainable layer-specific weight matrix; $\mathbf{H}^{(l)}\in\mathbb{R}^{N\times D}$ is the output of the previous \textsc{Gcn} layer where $N$ is number of entities and $D$ is the feature dimension; $\mathbf{H}^{(0)}$ is randomly initialised. We use the output of the last \textsc{Gcn} layer as the graph structure embedding $\mathbf{F}_G$.


\stitle{Visual embedding.}
We use \textsc{ResNet}-152 \citep{he2016deep}, pre-trained on the ImageNet \citep{deng2009imagenet} recognition task, as the feature extractor for all images. For each image, we do a forward pass and take the last layer's output before logits as the image representation (the \textsc{ResNet} itself is not fine-tuned). The feature is sent through a trainable feed-forward layer for the final image embedding:
\begin{equation}
    \mathbf{F}_I = \mathbf{W}_I \cdot \textsc{ResNet}(\mathcal{I}) + \mathbf{b}_I.
\end{equation}
The CNN-extracted visual representation is expected to capture both low-level similarity and high-level semantic relatedness between images \citep{kiela2014learning}.\footnote{We compared several popular pre-trained visual encoders but found no substantial difference (\Cref{sec:ablation_study}).}

\stitle{Relation and attribute embeddings.}
\citet{yang2019aligning} showed that modelling relations and attributes with \textsc{Gcn}s could pollute entity representations due to noise from neighbours. Following 
their investigation, we adopt a simple feed-forward network for mapping relation and attribute features into low-dimensional spaces:
\begin{equation}
\begin{split}
    \mathbf{F}_R = \mathbf{W}_R \cdot \mathbf{R} + \mathbf{b}_R;\\
    \mathbf{F}_A = \mathbf{W}_A \cdot \mathbf{A} + \mathbf{b}_A.\\
\end{split}
\end{equation}

\stitle{Modality fusion.}
We first $l_2$-normalise each feature matrix by row and then fuse multi-modal features by trainable weighted concatenation:
\begin{equation}
    \mathbf{F}_J = \bigoplus_{i=1}^n \Bigg[ \frac{e^{w_i}}{\sum_{j=1}^n e^{w_j}} \cdot \mathbf{F}_i\Bigg] ,
    \label{eq:fusion}
\end{equation}
where $n$ is the number of modalities; $w_i$ is an attention weight 
for the $i$-th modality. They are sent to a softmax before being multiplied to each modality's $l_2$-normalised representation, ensuring that the normalised weights sum to $1$.

\subsection{Alignment Learning}
On top of the multi-modal embeddings $\mathbf{F}_J$ for all entities, we compute the similarity of all bi-graph entity pairs and align them using an NCA loss. The training set is expanded using iterative learning.

\stitle{Embedding similarity.}
Let $\mathbf{F}_J^s, \mathbf{F}_J^t$ denote embeddings of the source and target entities $\mathcal{E}_s$ and $\mathcal{E}_t$ respectively. We compute their cosine similarity matrix
$\mathbf{S} = \langle \mathbf{F}_J^s, \mathbf{F}_J^t\rangle\in\mathbb{R}^{|\mathcal{E}_s|\times |\mathcal{E}_t|}$, where each entry $\mathbf{S}_{ij}$ corresponds to the cosine similarity between the $i$-th entity in $\mathcal{E}_s$ and the $j$-th in $\mathcal{E}_t$.

\stitle{NCA loss.}
\label{sec:loss}
Inspired by 
the NCA-based text-image matching approach proposed by \citet{liu2020hal},
we adopt an NCA loss of a similar form. It uses both local and global statistics to measure importance of samples and punishes hard negatives with a soft weighting scheme. 
This seeks to mitigate the \emph{hubness problem} \citep{radovanovic2010hubs}
in 
an embedding space. The loss is formulated below:
\begin{equation}
    \begin{split}
\mathcal{L} = \frac{1}{N}\sum_{i=1}^{N} \Bigg(  \frac{1}{\alpha}\log\Big(1+\sum_{m\not = i} e^{\alpha \mathbf{S}_{mi}}\Big) +  \\
\frac{1}{\alpha}\log\Big(1+\sum_{n\not=i} e^{\alpha \mathbf{S}_{in} }\Big) - \log \Big(1+\beta  \mathbf{S}_{ii} \Big)\Bigg) ,
\label{eq:loss}
\end{split}
\end{equation}
where $\alpha,\beta$ are temperature scales; $N$ is the number of 
pivots within the mini-batch. We apply such loss on each modality separately and also on the merged multi-modal representation as specified in \cref{eq:fusion}. The joint loss is written as:
\begin{equation}
\mathcal{L}_{\text{Joint}} = \sum_i^{n} \mathcal{L}_{i}+ \mathcal{L}_{\text{Multi-modal}}
\end{equation}
where 
$\mathcal{L}_i$ represents the loss term for aligning the $i$-th modality; $\mathcal{L}_{\text{Multi-modal}}$ is applied on the merged representation $\mathbf{F}_J$ and is used for training the modality weights only. The reason for having separate terms for different modalities is that we use different hyper-parameters to accommodate their drastically distinct feature distributions. For all terms we used $\beta=10$, but we picked different $\alpha$'s: $\alpha=5$ for $\mathcal{L}_{G}$; $\alpha=15$ for $\mathcal{L}_{R}, \mathcal{L}_{A},\mathcal{L}_{I},\mathcal{L}_{J}$.

\stitle{Iterative learning.}
\label{sec:iter}
To improve learning with very few training pivots,
we incorporate an iterative learning (\textsc{il}) strategy to propose more pivots from unaligned entities.
In contrast to previous work
\citep{sun2018bootstrapping}, we add a probation technique. In detail, for every $K_e$ epochs, we make a new round of proposal. Each pair of cross-graph entities that are mutual nearest neighbours is proposed and added into a candidate list. If a proposed entity pair remains mutual nearest neighbours throughout $K_s$ consecutive rounds (i.e. the probation phase), we permanently add it into the training set. Therefore, the candidate list refreshes every $K_e \cdot K_s$ epochs.
In practice, we find that the probation technique has made the pivot discovery process more stable.

\begin{algorithm}[t!]
\small
\SetAlgoLined
\SetKwInOut{Input}{input}
\SetKwInOut{Output}{output}
\Input{visual embeddings from entities in the two graphs ($\mathbf{F}_I^1$, $\mathbf{F}_I^2$); pivot dictionary size ($n$)} 
\Output{pivot dictionary}
$\mathbf{M} \leftarrow \large\langle \mathbf{F}_I^1$, $\mathbf{F}_I^2 \large\rangle $ \algorithmiccomment{get similarity matrix} \\
$\mathbf{m}_s \leftarrow \texttt{sort}(\mathbf{M})$ \algorithmiccomment{sort elements of $\mathbf{M}$} \\
$\mathcal{S}\leftarrow \{\}$ \algorithmiccomment{initialise seed dictionary} \\
$\mathcal{R}_u \leftarrow \{\}; \mathcal{C}_u \leftarrow \{\}$ \algorithmiccomment{for recording used row/column} \\
\While{$|\mathcal{S}| != n$}{
    $m \leftarrow \mathbf{m}_s\texttt{.pop()}$ \algorithmiccomment{get the highest ranked score} \\
    \If{\emph{$m\texttt{.ri}\not\in\mathcal{R}_u\ \&\  m\texttt{.ci}\not\in\mathcal{C}_u$}}{ 
    $\mathcal{S}\leftarrow \mathcal{S}\cup (m\texttt{.ri},m.\texttt{ci})$  \algorithmiccomment{store the pair} \\
    $\mathcal{R}_u\leftarrow \mathcal{R}_u\cup m\texttt{.ci}$ \\ $\mathcal{C}_u\leftarrow \mathcal{C}_u\cup m\texttt{.ri}$ 
    }  
} 
\Return $\mathcal{S}$ \algorithmiccomment{return the obtained visual pivot dictionary}
 \caption{Visual pivot induction.}
 \label{alg:unsup}
\end{algorithm}

\subsection{Unsupervised Visual Pivoting}\label{sec:method_unsup}

 

Previous EA methods require annotated pivots that may not be widely available across KGs \cite{zhuang2017hike,chen2017multigraph}. Our method, however, can naturally extend to an unsupervised setting where visual similarities are leveraged to infer correspondence between KGs, and no annotated cross-graph pivots are required. 
All cross-graph supervision comes from an automatically induced visual dictionary (visual pivots) containing the most visually alike cross-graph entities. Specifically, we first compute cosine similarities of all cross-graph entities' visual representations in the data set. Then we sort the cosine similarity matrix from high to low. We collect visual pivots starting from the most similar pairs. Once a pair of entities is collected, all other links associated with the two entities are discarded. In the end, we obtain a cross-graph pivot list that records the top-$k$ visually similar entity pairs without repetition of entities. From these visual pivots, we apply iterative learning (\Cref{sec:iter}) to expand the training set. The algorithm of obtaining visual pivots is formally described in \Cref{alg:unsup}. Let $n$ be the number of entities in one language, the algorithm takes $\mathcal{O}(n^2\log(n^2) + n^2) = \mathcal{O}(n^2\log n)$. 

\begin{table*}[!ht]
\small
\setlength{\tabcolsep}{4pt}
\renewcommand{\arraystretch}{0.9}
\caption{\small{Cross-lingual EA results on DBP15k. Comparison with related works with and without using \textsc{il}. ``-'' means not reported by the original paper. ``$\ast$'' indicates our reproduced results for which any use of machine translation or cross-lingual alignment labels other than those provided in the benchmark are removed. \textbf{Bold} numbers are the best models and \underline{\textbf{underline}} marks statistical significance ($p$-value$<0.05$)}.}
\label{tab:dbp15k}
\centering
\begin{tabular}{clccccccccccccccccccccccccccccc}
\toprule
 & \multirow{2}{*}{model} &     \multicolumn{3}{c}{FR$\rightarrow$EN} & $\ $ &  \multicolumn{3}{c}{JA$\rightarrow$EN} &$\ $ &  \multicolumn{3}{c}{ZH$\rightarrow$EN} \\
 \cmidrule{3-5}\cmidrule{7-9}\cmidrule{11-13}
  & &  \scriptsize H@1 & \scriptsize H@10 &  \scriptsize MRR & & \scriptsize H@1 & \scriptsize H@10 & \scriptsize MRR & & \scriptsize H@1 & \scriptsize H@10 &  \scriptsize MRR \\
\midrule 
 \multirow{11}{*}{\rotatebox[origin=c]{90}{\textsc{w/o il}}} & \textsc{MTransE} \citep{chen2017multigraph} & .224 & .556 & .335 & & .279 & .575 & .349  & & .308 & .614 & .364 \\
  & \textsc{Jape} \citep{sun2017cross} & .324 & .667 & .430 & & .363 & .685 & .476 & & .412 & .745 & .490 \\
  & \textsc{Gcn} \citep{wang2018cross} &  .373 & .745 & .532 & & .399 & .745 & .546 & & .413 & .744 & .549\\
  & \textsc{MuGnn} \citep{cao2019multi} & .495 & .870 & .621 & & .501 & .857 & .621 & & .494 & .844 & .611\\
 & \textsc{Rsn} \citep{guo2019learning} & .516 & .768 & .605 & & .507 & .737 & .590 & & .508 & .745 & .591 \\
  & \textsc{Kecg} \citep{li2019semi} & .486 & .851 & .610 & &  .490  & .844 &.610  & &.478 & .835 & .598  \\
 & \textsc{Hman} \citep{yang2019aligning} & .543 & .867 & - & & .565 & .866 & - & & .537 & .834 & -\\
 & \textsc{Gcn-Je} \citep{wu2019jointly} & .483 & .778 & - & & .466 & .746 & - & & .459 & .729 & - \\
 & \textsc{Gmn} \citep{xu2019cross}$^\ast$ & .596 & .876 & .679 & & .465 & .728 & .580 & & .433 & .681 & .479\\
  &\textsc{AliNet} \citep{sun2020alinet} & .552 &  .852 & .657 & & .549 & .831 & .645 & & .539 & .826 & .628\\

 &  \CC{20} &  \CC{20}\underline{\textbf{.715}} &  \CC{20}\underline{\textbf{.936}} &  \CC{20}\underline{\textbf{.795}}  & \CC{20} &  \CC{20}\underline{\textbf{.716}} &  \CC{20}\underline{\textbf{.926}} &  \CC{20}\underline{\textbf{.792}} & \CC{20} &  \CC{20}\underline{\textbf{.720}} &  \CC{20}\underline{\textbf{.925}} &  \CC{20}\underline{\textbf{.793}}  \\
  & \multirow{-2}{*}{\CC{20}\modelname \textsc{w/o il}} & \CC{20} \scriptsize $\pm.003$ & \CC{20} \scriptsize $\pm.002$ & \CC{20} \scriptsize $\pm.004$  & \CC{20}  & \CC{20}\scriptsize $\pm.008$  & \CC{20}\scriptsize $\pm.004$  & \CC{20} \scriptsize $\pm.006$ & \CC{20}  & \CC{20}\scriptsize $\pm.004$  & \CC{20}\scriptsize $\pm.006$  & \CC{20} \scriptsize $\pm.003$ \\
\midrule
\multirow{5}{*}{\rotatebox[origin=c]{90}{\textsc{w/ il}}} & \textsc{BootEA} \citep{sun2018bootstrapping} & .653 & .874 & .731 & & .622 & .854 & .701 & & .629 & .847 & .703\\
& \textsc{Mmea} \citep{shi2019modeling} & .635 & .878 & - & & .623 & .847 & - & & .647 & .858 & - \\
& \textsc{Naea} \citep{zhu2019neighborhood} & .673 & .894 & .752 & & .641 & .873 & .718 & & .650 & .867 & .720 \\
& \CC{20} & \CC{20} \underline{\textbf{.793}} & \CC{20}\underline{\textbf{.942}} & \CC{20}\underline{\textbf{.847}} & \CC{20}  & \CC{20}\underline{\textbf{.762}} & \CC{20}\underline{\textbf{.913}} & \CC{20}\underline{\textbf{.817}} & \CC{20} & \CC{20}\underline{\textbf{.761}} & \CC{20}\underline{\textbf{.907}} & \CC{20}\underline{\textbf{.814}} \\
 & \multirow{-2}{*}{\CC{20}\modelname \textsc{w/ il}} & \CC{20} \scriptsize $\pm.005$ & \CC{20} \scriptsize $\pm.002$ & \CC{20} \scriptsize $\pm.004$  & \CC{20}  & \CC{20}\scriptsize $\pm.008$  & \CC{20}\scriptsize $\pm.003$  & \CC{20} \scriptsize $\pm.006$ & \CC{20}  & \CC{20}\scriptsize $\pm.008$  & \CC{20}\scriptsize $\pm.005$  & \CC{20} \scriptsize $\pm.006$ \\
 \bottomrule
\end{tabular}
\end{table*}

Our approach is related to some recent efforts on word translation with images \citep{bergsma2011learning,kiela2015visual,hewitt2018learning}. However, those efforts focus on obtaining cross-lingual parallel signals from web-crawled images provided by search engines (e.g. Google Image Search). This can result in noisy data caused by issues like ambiguity in text. For example, for a query \emph{mouse}, the search engine might return images for both the rodent and the computer mouse. 
Visual pivoting is thus more suitable in the context of EA, as images provided by KGs are mostly human-verified and disambiguated, serving as gold visual representations of the entities. Moreover, cross-graph entities are not necessarily cross-lingual, meaning that the technique could also benefit a monolingual scenario.

\section{Experiments}
\label{sec:exp}


In this section, we conduct experiments on two benchmark data sets (\Cref{sec:exp_setting}), under both semi- and unsupervised settings (\Cref{sec:exp_unsup}).
We also provide detailed ablation studies on different model components (\Cref{sec:ablation_study}),
and study the impact of incorporating visual representations on long-tail entities (\Cref{sec:long_tailed}).

\begin{figure*}[!t]
    \centering
    \includegraphics[width=.33\linewidth]{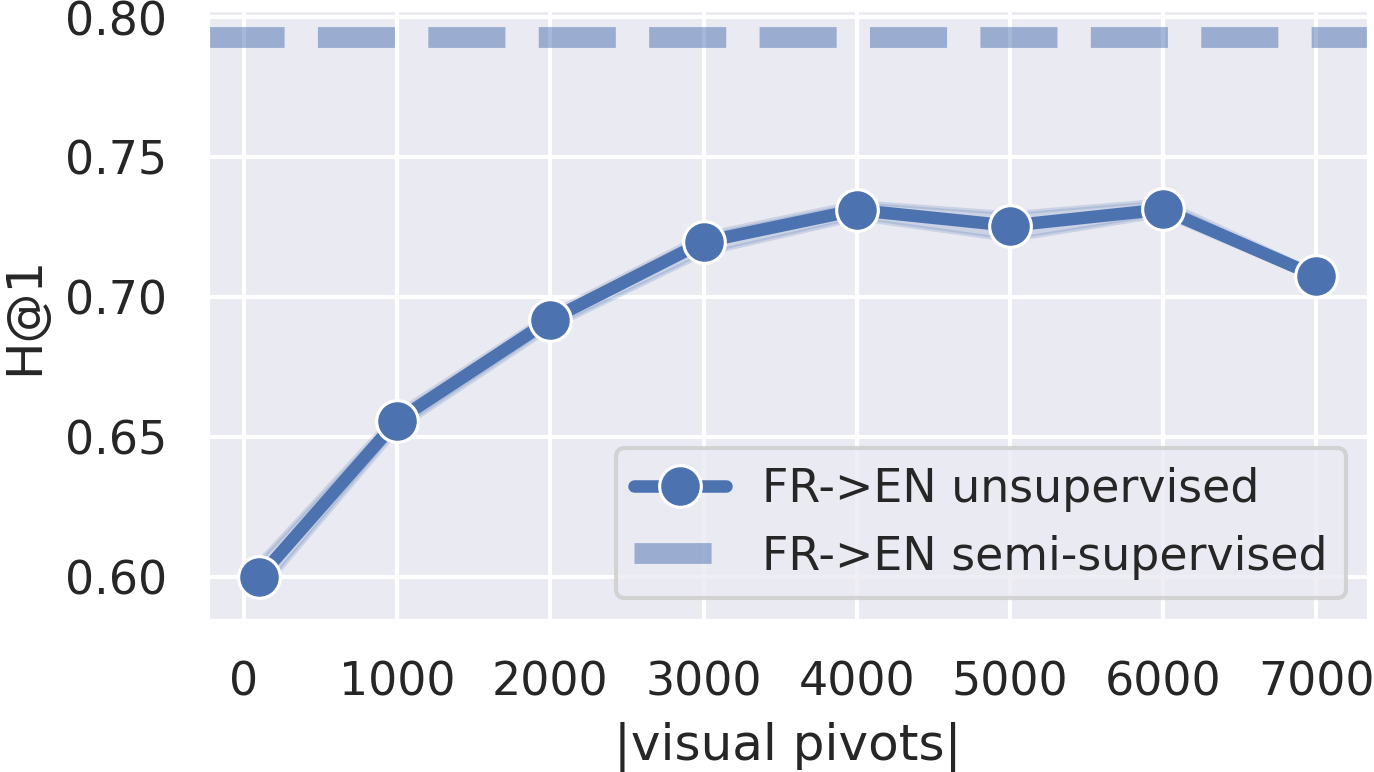}
    \includegraphics[width=.33\linewidth]{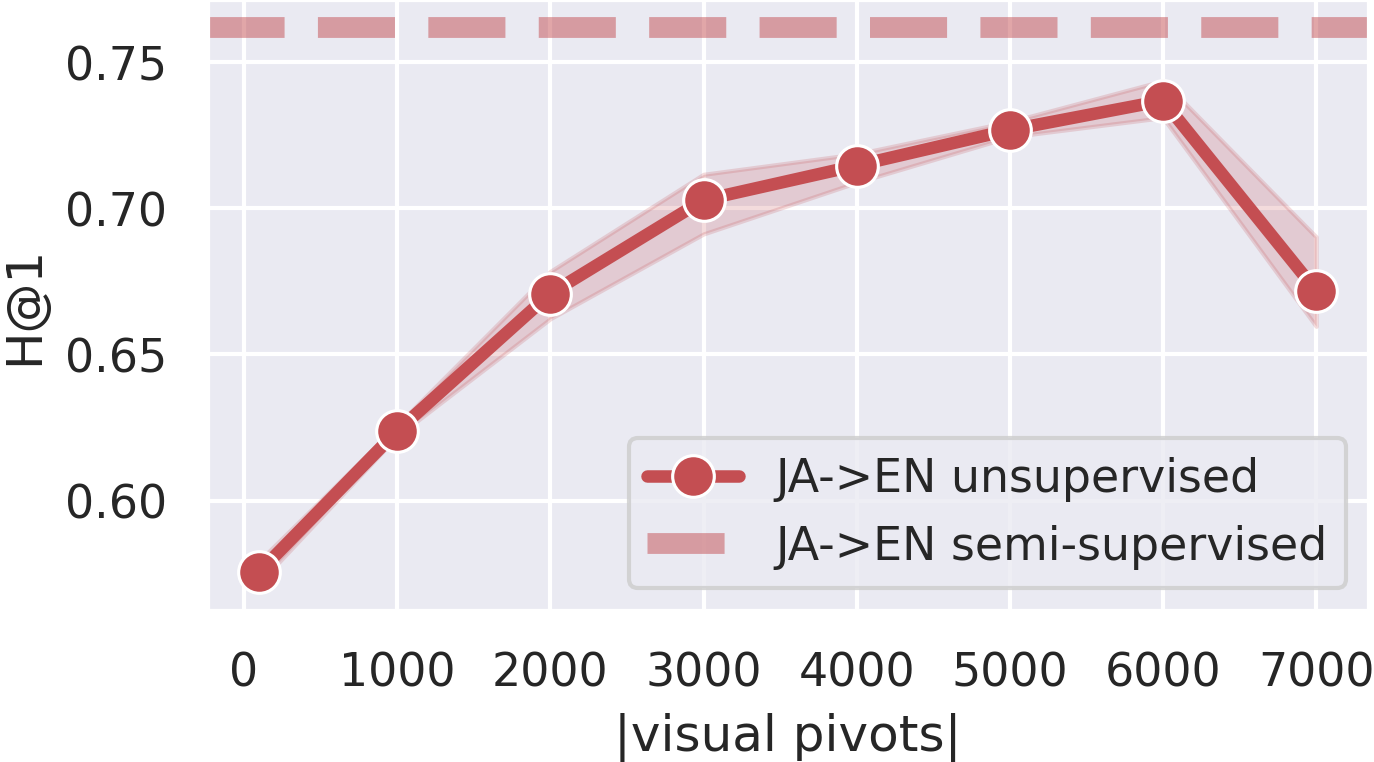}
    \includegraphics[width=.33\linewidth]{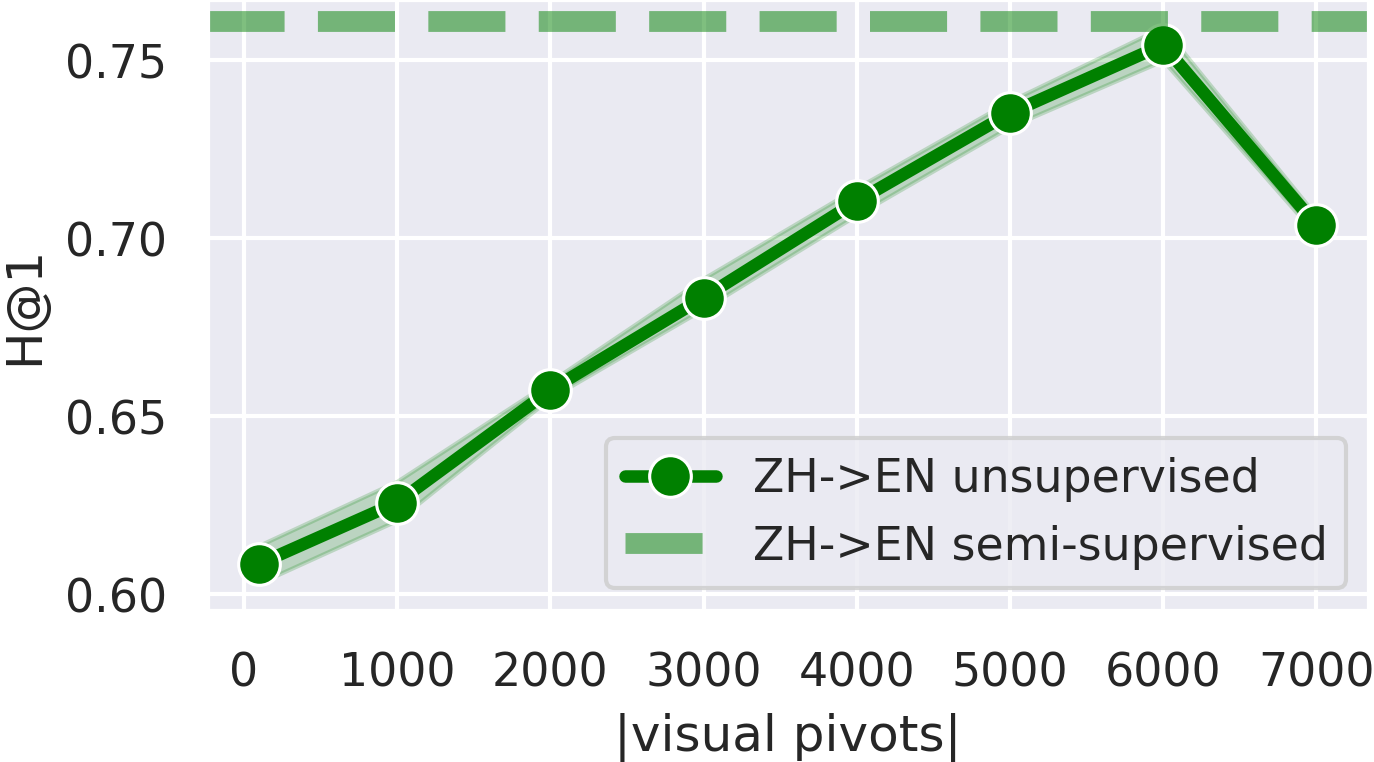}
    \caption{\small{Unsupervised \modelname vs. semi-supervised \modelname. Plotting H@1 against number of induced visual seeds for jump-starting the training.}}
    \label{fig:unsup}
\end{figure*}

\subsection{Experimental Settings}
\label{sec:exp_setting}
\stitle{Data sets.} 
The experiments are conducted on DBP15k \citep{sun2017cross} and DWY15k \citep{guo2019learning}. DBP15k is a widely used cross-lingual EA benchmark. It contains four language-specific KGs from DBpedia, and has three bilingual EA settings, i.e., French-English (FR-EN), Japanese-English (JA-EN) and Chinese-English (ZH-EN). 
DBpedia has also released images for English, French and Japanese versions. 
Note that since Chinese images are not released in DBpedia,
we extracted them from the raw Chinese Wikipedia dump with the same process as described by \citet{lehmann2015dbpedia}. DWY15k is a monolingual data set, focusing on EA for DBpedia-Wikidata and DBpedia-YAGO. It has two subsets, DWY15k-norm and DWY15k-dense, whereof the former is much sparser. As YAGO does not have image components, we experiment on DBpedia-Wikidata only. Note that not all but ca. 50-85\% entities have images, as shown in \Cref{Table:image_coverage_stat}. For an entity without an image, we assign a random vector sampled from a normal distribution, parameterised by the mean and standard deviation of other images. 
As for relation and attribute features, we extract them in the same way as \citet{yang2019aligning}.

\stitle{Model configurations.} The \textsc{Gcn} has two layers with input, hidden and output dimensions of 400, 400, 200 respectively. Attribute and relation features are mapped to 100-$d$. Images are transformed to 2048-$d$ features by \textsc{ResNet} and then mapped to 200-$d$.
For model variants without \textsc{il}, training is limited to 500 epochs. 
Otherwise, after the first 500 epochs, \textsc{il} is conducted for another 500 epochs with the configurations $K_e=5, K_s=10$ as described in \Cref{sec:iter}. We train all models using a batch size of 7,500. 
The models are optimised using AdamW \citep{loshchilov2018decoupled} with a learning rate of \texttt{5e-4} and a weight decay of \texttt{1e-2}.
More implementation details are available in \Cref{sec:implementation}~\cite{liu2021visual}.

\stitle{Evaluation protocols.} 
Following convention, we report three metrics on both data sets, including H@\{1,10\} (the proportion of ground truth being ranked no further than top \{1,10\}), and MRR (mean reciprocal rank). During inference, we use Cross-domain Similarity Local Scaling (\textsc{Csls}; \citeauthor{lample2018word}~\citeyear{lample2018word}) to post-process the cosine similarity matrix, which is employed by default in some recent works \citep{sun2019transedge,sun2020alinet}. $k=3$ is used for defining local neighbourhood of \textsc{Csls}. All models are run for 5 times with 5 different random seeds and the average with variances are reported. \textbf{Bold} numbers in tables come from the best models and \underline{\textbf{underline}} means with statistical significance (p-value $<0.05$ in t-test).

The baseline results on DBP15k come from ten methods with \textsc{il}, and three without. 
We accordingly report the results by \modelname with and without \textsc{il}.
Note that a few methods may incorporate extra cross-lingual alignment labels by initialising training with machine translation \citep{wu2019jointly,yang2019aligning} or pre-aligned word vectors \citep{xu2019cross}. For fair comparison in this study, we report results from the versions of these methods that do not use any alignment signals apart from the training data. 
On DWY15k, there are also two settings in the literature, different in whether to use the surface form embeddings of monolingual entities \citep{yang2020cotsae} or not \citep{guo2019learning}. 
We report results from \modelname with and without using surface forms\footnote{When incorporating surface form, we use \textsc{FastText} \citep{bojanowski2017enriching} to embed surface strings into low-dimensional vectors $\mathbf{S}\in\mathbb{R}^{N \times d_s}$ ($d_s=300$), and learn a linear transformation to obtain final representations in 100-$d$: $\mathbf{F}_S = \mathbf{W}_S\cdot \mathbf{S} + \mathbf{b}_S$. We merge and train the surface form modality in the same way as the other modalities.}, and compare with five SOTA baselines. 

\subsection{Main Results}\label{sec:main_results}

\stitle{Semi-supervised EA.}
\label{sec:exp_semi}
\Cref{tab:dbp15k} reports the results on semi-supervised cross-lingual EA. 
This setting compares \modelname 
with baseline methods using the original data split of DBP15k, i.e., using 30\% of the EA labels for training.
Consequently, \modelname achieves SOTA performance and surpasses baseline models drastically both with or without \textsc{il}. 
Specifically, in the \textsc{w/o il} setting, \modelname leads to 12.3-17.6\% absolute improvement in H@1 over the best baseline.
When incorporating \textsc{il}, \modelname gains 11.9-12.5\% absolute improvement in H@1 over the best \textsc{il}-based baseline method.
This indicates that incorporating the visual representation competently improves the cross-lingual entity representations for inferring their correspondences, without the need of additional supervision labels.
The results on monolingual EA generally exhibit similar observations.
As reported in \Cref{tab:dwy15k},
without incorporating surface form information, \modelname surpasses the strongest baseline method by 16.8\% in H@1 on the normal split and 4.2\% on the dense split. 
With surface forms considered, \modelname offers near-perfect results, outperforming the SOTA method by 26.8\% in H@1 on the normal split and 6.6\% in H@1 on the dense split. 
The experiments here indicate that, \modelname is able to substantially improve SOTA EA systems under both monolingual and cross-lingual settings.

\begin{table}[!t]
\small
\setlength{\tabcolsep}{0.5pt}
\renewcommand{\arraystretch}{0.9}
\caption{\small{Monolingual EA results on DWY15k-DW (N: normal split; D: dense split). \modelname using \textsc{il} is compared with related works with and without using surface forms (\textsc{w/ sf} \& \textsc{w/o sf}).}}
\label{tab:dwy15k}
\centering
\begin{tabular}{clcccccccccccccccccc}
\toprule
 & \multirow{2}{*}{model} &     \multicolumn{3}{c}{\scriptsize  DBP$\rightarrow$WD (N)} & &  \multicolumn{3}{c}{\scriptsize  DBP$\rightarrow$WD (D)} \\
 \cmidrule{3-5}\cmidrule{7-9}
  & &  \scriptsize H@1 & \scriptsize H@10 & \scriptsize MRR & & \scriptsize H@1 & \scriptsize H@10 & \scriptsize MRR \\
\midrule 
  \multirow{7}{*}{\rotatebox[origin=c]{90}{\textsc{w/o sf}}} &
 \textsc{BootEA} \scriptsize \citep{sun2018bootstrapping}& .323 & .631 &.420  & & .678 & .912 & .760\\
  & \textsc{Gcn} \scriptsize \citep{wang2018cross}& .177 & .378  & .250 & & .431 & .713 & .530 \\
  & \textsc{Jape} \scriptsize \citep{sun2017cross} & .219 & .501 & .310 & $\ $ & .393 & .705 & .500\\
  & \textsc{Rsn} \scriptsize \citep{guo2019learning} &  .388 & .657 & .490 & & .763 & .924 & .830\\
  & \textsc{Cotsae} \scriptsize \citep{yang2020cotsae} & .423 & .703 & .510 & & .823 & .954 & .870\\
  & \CC{20}  &  \CC{20}\underline{\textbf{.593}} &  \CC{20}\underline{\textbf{.775}}  &  \CC{20}\underline{\textbf{.655}} & \CC{20} & \CC{20}\underline{\textbf{.874}} &  \CC{20}\underline{\textbf{.962}} &  \CC{20}\underline{\textbf{.908}} \\
  & \multirow{-2}{*}{\CC{20}\modelname \textsc{w/o sf}} & \CC{20} \scriptsize $\pm.004$ & \CC{20} \scriptsize $\pm.005$ & \CC{20} \scriptsize $\pm.003$  & \CC{20}  & \CC{20}\scriptsize $\pm.002$  & \CC{20}\scriptsize $\pm.003$  & \CC{20} \scriptsize $\pm.002$ \\
\midrule
 \multirow{3}{*}{\rotatebox[origin=c]{90}{\textsc{w/ sf}}} & \textsc{Cotsae} \scriptsize\citep{yang2020cotsae} & .709 & .904 & .770 & & .922 & .983 & .940 \\
 &  \CC{20} &  \CC{20}\underline{\textbf{.985} }&  \CC{20}\underline{\textbf{.995}} &  \CC{20}\underline{\textbf{.989}} & \CC{20} &  \CC{20}\underline{\textbf{.994}} &  \CC{20}\underline{\textbf{1.0}} &  \CC{20}\underline{\textbf{.996}}\\
  & \multirow{-2}{*}{\CC{20}\modelname \textsc{w/ sf}} & \CC{20} \scriptsize $\pm.001$ & \CC{20} \scriptsize $\pm.000$ & \CC{20} \scriptsize $\pm.001$  & \CC{20}  & \CC{20}\scriptsize $\pm.001$  & \CC{20}\scriptsize $\pm.001$  & \CC{20} \scriptsize $\pm.000$ \\
\bottomrule
\end{tabular}
\end{table}

\begin{table}[!t]
\small
\setlength{\tabcolsep}{0.8pt}
\renewcommand{\arraystretch}{0.9}
\caption{\small{Comparing unsupervised and semi-supervised \modelname results on DBP15k.}}
\label{tab:unsupervised_dbp15k}
\centering
\begin{tabular}{clccccccccccccccccccccccccccccc}
\toprule
 \multirow{2}{*}{setting}  &  \multicolumn{3}{c}{FR$\rightarrow$EN} & $\ $ & \multicolumn{3}{c}{JA$\rightarrow$EN} & $\ $ &  \multicolumn{3}{c}{ZH$\rightarrow$EN} \\
 \cmidrule{2-4}\cmidrule{6-8} \cmidrule{10-12}
   &  {\scriptsize H@1} &\scriptsize H@10 &\scriptsize MRR & &\scriptsize H@1 &\scriptsize H@10 &\scriptsize MRR & &\scriptsize H@1 &\scriptsize H@10 &\scriptsize MRR \\
  \midrule
  \multirow{2}{*}{\scriptsize Unsup.}  & .731 & .909 & .792 & & .737 & .890 & .791 & & .752 & .895 & .804 \\ 
 & \scriptsize $\pm.004$ & \scriptsize $\pm.003$ & \scriptsize $\pm.003$  & & \scriptsize $\pm.008$  & \scriptsize $\pm.004$  & \scriptsize $\pm.006$ &  & \scriptsize $\pm.006$  & \scriptsize $\pm.004$  & \scriptsize $\pm.005$ \\
\midrule
 \multirow{2}{*}{\scriptsize Semi-sup.}  & .793 & .942 & .847 & & .762 & .913 & .817   & & .761 & .907 & .814  \\
 & \scriptsize $\pm.003$ & \scriptsize $\pm.002$ & \scriptsize $\pm.004$  & & \scriptsize $\pm.008$  & \scriptsize $\pm.004$  & \scriptsize $\pm.006$ &  & \scriptsize $\pm.004$  & \scriptsize $\pm.006$  & \scriptsize $\pm.003$ \\
\bottomrule
\end{tabular}
\end{table}

\stitle{Unsupervised EA.}\label{sec:exp_unsup}
We also use the visual pivoting technique (\Cref{sec:method_unsup}) with \modelname to conduct unsupervised EA, without using any annotated alignment labels. 
We compare \modelname's best unsupervised and semi-supervised results in \Cref{tab:unsupervised_dbp15k}. 
The unsupervised \modelname yields 1.9-6.3\% lower H@1 than the semi-supervised version,
but still notably outperforms the best semi-supervised baseline (\Cref{tab:dbp15k}) by 5.6-10.1\%.  We change the number of visual seeds by adjusting the threshold $n$ in \Cref{alg:unsup}, and test the model's sensitiveness to the threshold. As shown in \Cref{fig:unsup},
the optimal seed size is 4k on FR$\rightarrow$EN and 6k on JA$\rightarrow$EN and ZH$\rightarrow$EN. 

It is worth noticing that a good alignment (H@1$>$55\%) can be obtained using as few as a hundred visual seeds. As the number of seeds grows, the model gradually improves, reaching $>$70\% H@1 with more than 3k seeds. Then the scores plateau for a period and start to decrease with more than 4k (on FR$\rightarrow$EN) or 6k (JA$\rightarrow$EN, ZH$\rightarrow$EN) seeds. This is because a large visual seed dictionary starts to introduce noise.
Empirically, we find that a 0.85 cosine similarity threshold is a good cut-off point.

\subsection{Ablation Study}
\label{sec:ablation_study}

We report an ablation study of \modelname in \Cref{tab:ablation_study} using DBP15k (FR$\rightarrow$EN).
As shown, \textsc{il} brings ca. 8\% absolute improvement.
This gap is smaller than what has been reported previously \citep{sun2018bootstrapping}. 
This is because the extra visual supervision in our method already allows the model to capture fairly good alignment in the first 500 epochs, leaving smaller room for further improvement from \textsc{il}. 
\textsc{Csls} gives minor but consistent improvement to all metrics during inference. While \textsc{Csls} is mainly used to reduce hubs in a dense space such as textual embeddings \citep{lample2018word}, we suspect that it cannot bring substantial improvement to our sparse multi-modal space. Besides, the hubness problem is already partly tackled by our NCA loss.
The sparseness of multi-modal space can also explain our choice of $k=3$, which we found to be better than the previous $k=10$. 

Regarding the impact from different modalities, structure remains the most important for our model. 
Dropping structural embedding decreases H@1 from ca. 80\% to below 40\%, cutting the performance by half. This is in line with the findings by \citet{yang2019aligning}. Image, attributes and relations are of similar importance. 
The removal of images and attributes decrease H@1 by 4-5\% while
removing relations causes ca. 3\% drop in H@1. This general pattern roughly corresponds to
the modality attention weights. On DBP15k, while all weights start at 0.25, 
after training, they become ca. 0.45, 0.21, 0.17 and 0.16 for structures, images, relations and attributes respectively.
 

\begin{table}[!t]
\small
\renewcommand{\arraystretch}{0.9}
\caption{\small{Ablation study of \modelname based on DBP15k (FR$\rightarrow$EN).}}
\label{tab:ablation_study}
\centering
\begin{tabular}{lcccccccccccccccccc}
\toprule
model & \scriptsize  H@1 & \scriptsize H@10 & \scriptsize MRR \\
\midrule
\textsc{w/o} structure & .391 \scriptsize$\pm.004$ & .514 \scriptsize$\pm.003$ & .423 \scriptsize$\pm.004$ \\
\textsc{w/o} image & .749 \scriptsize$\pm.002$ & .929 \scriptsize$\pm.002$ & .817 \scriptsize$\pm.001$ \\
\textsc{w/o} attribute & .750 \scriptsize$\pm.003$ &.927 \scriptsize$\pm.001$ & .813 \scriptsize$\pm.003$ \\
\textsc{w/o} relation & .763 \scriptsize$\pm.006$ & .928 \scriptsize$\pm.003$ & .823 \scriptsize$\pm.004$  \\
\midrule 
  \textsc{w/o il}  & .715 \scriptsize$\pm.003$ & .936 \scriptsize$\pm.002$ & .795 \scriptsize$\pm.004$ \\
 \textsc{w/o Csls} & .786 \scriptsize$\pm.005$ & .928 \scriptsize$\pm.001$ & .838 \scriptsize$\pm.003$ \\
\midrule
full model & \underline{\textbf{\textbf{.793}}} \scriptsize$\pm.003$ & \underline{\textbf{\textbf{.942}}} \scriptsize$\pm.002$ & \underline{\textbf{\textbf{.847}}} \scriptsize$\pm.004$ \\
\bottomrule
\end{tabular}
\end{table}



In addition, we explore several other popular options of pre-trained visual encoder architectures including ResNet \citep{he2016deep}, GoogLeNet \citep{szegedy2015going}, DenseNet \citep{huang2017densely} and Inception v3 \citep{szegedy2016rethinking} as feature extractors for images. One of the variants, ResNet (Places365) \citep{zhou2017places}, is pre-trained on a data set from the outdoor-scene domain and is expected to be better at capturing location-related information. In general, we found little difference in model performance with different visual encoders. As suggested in \Cref{Table:visual_encoder}, variances from different models across different metrics are generally $<1\%$. All numbers reported in the paper have been using ResNet152 as it is one of the most widely used visual feature extractor. It is also possible to fine-tune the visual encoder with the full model in an end-to-end fashion. However, the computation cost would be extremely large under our setting.

\begin{table}[t!] 
\small
\renewcommand{\arraystretch}{0.9}
\setlength{\tabcolsep}{3.2pt}
\caption{\small Comparing visual encoders. Results are obtained on DBP15k (FR$\rightarrow$EN) using \modelname \textsc{w/o il}.} 
\label{Table:visual_encoder}
\centering
\begin{tabular}{lcccc}
\toprule
visual encoder & H@1 & H@10 & MRR \\
\midrule
ResNet50	& .713 \scriptsize $\pm.003$ & \textbf{.938} \scriptsize $\pm.002$ & .794 \scriptsize $\pm.003$ \\
ResNet50 (places365) & .710 \scriptsize $\pm.002$	& .937 \scriptsize $\pm.002$ & .792 \scriptsize $\pm.002$ \\
ResNet152	& .715 \scriptsize $\pm.003$	 & .936 \scriptsize  $\pm.002$	& .795 \scriptsize  $\pm.004$ \\
DenseNet201	& \textbf{.716} \scriptsize $\pm.005$	& .935 \scriptsize $\pm.002$	& \textbf{.796} \scriptsize $\pm.003$ \\
Inception v3 & .711 \scriptsize $\pm.002$ & .936 \scriptsize $\pm.002$ & .792 \scriptsize $\pm.002$ \\
\bottomrule
\end{tabular}
\end{table}


\subsection{Analysis on Long-tail Entities}
\label{sec:long_tailed}

\begin{figure}[!t]
\centering
\small
\includegraphics[width=\linewidth]{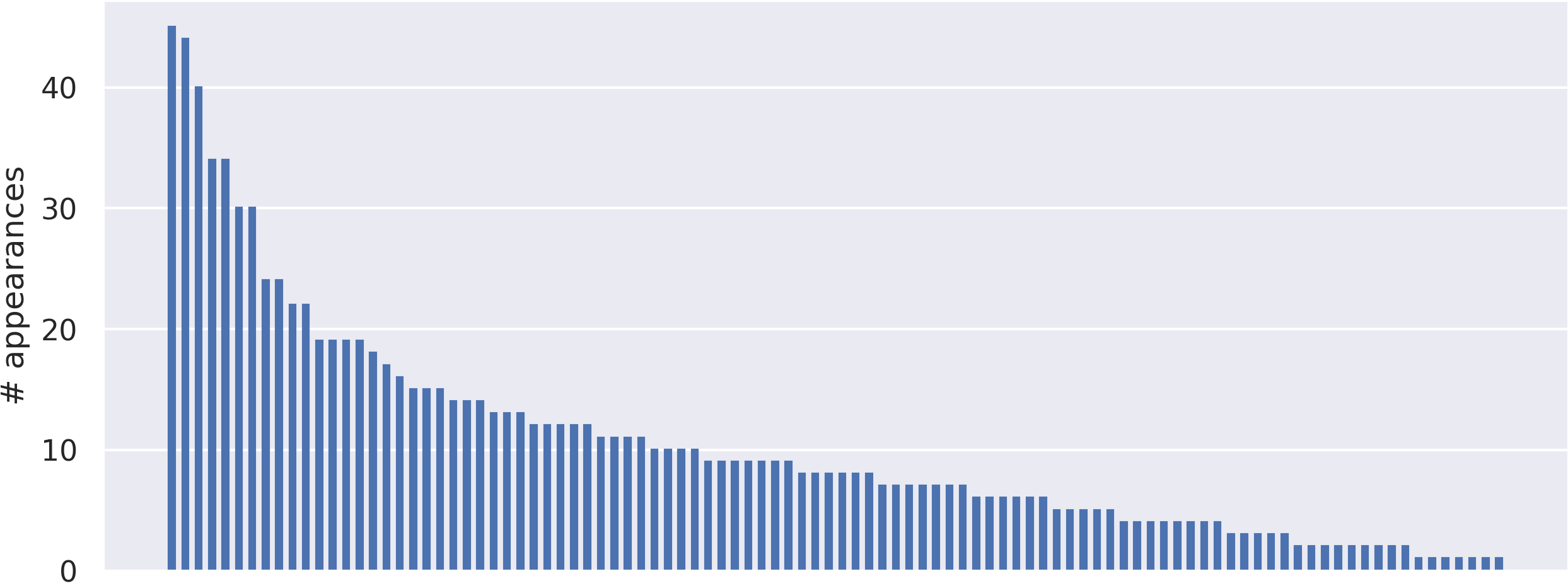}
 \caption{\small{Long-tailed distribution of entity appearances in KG triplets, using 100 randomly sampled entities in DBP15k (FR-EN).}}
 \label{fig:long_tail}
 \end{figure}

\begin{figure}[!t]
\centering

\includegraphics[width=0.95\linewidth]{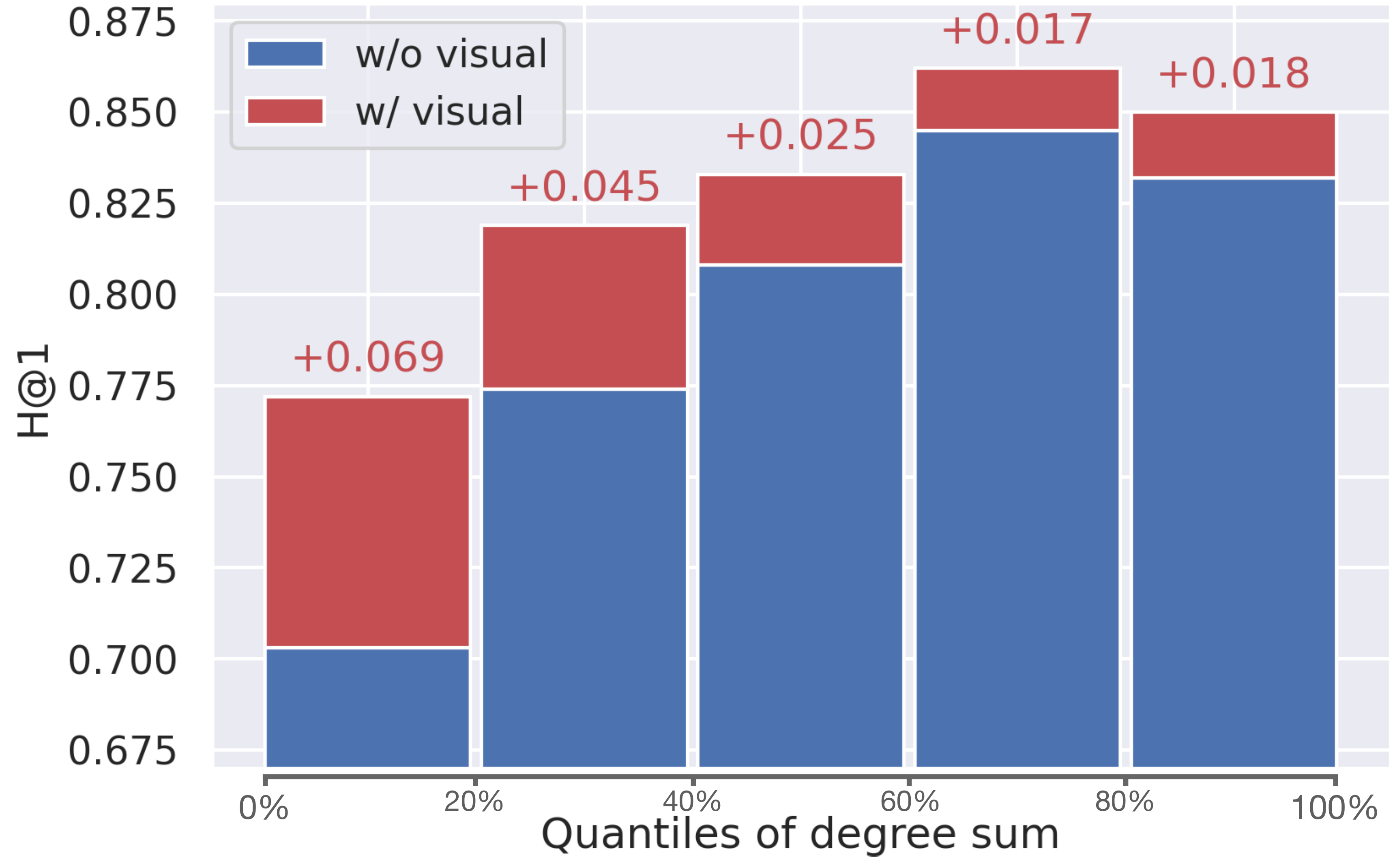}
\caption{\small{Plotting H@1 against different test splits on FR-EN (frequency low-to-high from left to right). Models w/ or w/o visual information are compared. The plot suggests that visual information has improved long-tail entities' alignment more.}}
\label{fig:long_tail_five_splits}
 \end{figure}

Like lexemes in natural languages, the occurrence of entities in KG triplets also follow a long-tailed distribution (\Cref{fig:long_tail}). Long-tail entities are poorly connected to others in the graph and thus have less structural information for inducing reliable representation and alignment.
We argue that images might remedy the issue by providing alternative sources of signal for representing these long-tail entities. 
To validate our hypothesis, we stratify the test set of DBP15k (FR$\rightarrow$EN) into five splits of entity pairs based on their degree centrality in the graphs. Specifically, for all entity pairs $(e_s,e_t)\in \mathcal{G}_s\times\mathcal{G}_t$ in the test set, we sort them by their degree sum, i.e., $\texttt{DegSum}(e_s,e_t) := \deg(e_s)+\deg(e_t)$, and split them into five sets of equal sizes, 
corresponding to five ranges of \texttt{DegSum} partitioned by 14, 18, 23 and 32, respectively. 
Across the five splits, we compare the performance by \modelname (\textsc{w/o il}) against its variant where visual inputs are disabled.
The results in \Cref{fig:long_tail_five_splits} suggest that entities in the lower ranges of degree centrality benefit more from the visual representations.
This demonstrates that the visual modality particularly enhances the match of long-tail entities which gain less information from other modalities.

 \begin{table*}[!ht] 
\small
\setlength{\tabcolsep}{4pt}
\renewcommand{\arraystretch}{0.9}
\caption{\small Image coverage statistics on DBP15k and DWY15k. The image coverage of DBP15k (ca. 65-85\%) is generally better than DWY15k (ca. 50-60\%).}
\label{Table:image_coverage_stat}
\centering
\begin{tabular}{ccccccccccccccc}
\toprule
  \multirow{2}{*}{} &  \multicolumn{2}{c}{FR$\leftrightarrow$EN} & & \multicolumn{2}{c}{JA$\leftrightarrow$EN} & & \multicolumn{2}{c}{ZH$\leftrightarrow$EN} & & \multicolumn{2}{c}{DBP$\leftrightarrow$WD (norm)} & & \multicolumn{2}{c}{DBP$\leftrightarrow$WD (dense)}\\
 \cmidrule{2-3}\cmidrule{5-6}\cmidrule{8-9}\cmidrule{11-12} \cmidrule{14-15} 
  & FR & EN & & JA & EN & & ZH & EN & &  DBP & WD && DBP & WD \\
  \midrule 
 image covered & 14,174 & 13,858 & & 12,739 & 13,741 & & 15,912 & 14,125 & & 8,517 & 8,791 & & 7,744 & 7,315\\
 all entities & 19,661 & 19,993 & & 19,814 & 19,780 & & 19,388 & 19,572 & & 15,000 & 15,000 & & 15,000 & 15,000 \\
\bottomrule
\end{tabular}
\end{table*}

\begin{figure}[t!]
  \centering
\begin{subfigure}[b]{\linewidth}
\centering
    \includegraphics[width=0.6\linewidth]{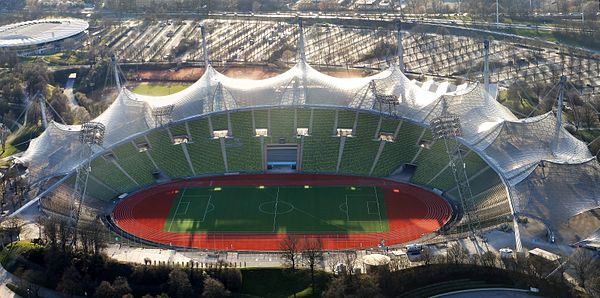}
    \caption{Query (French): \emph{Stade\_olympique\_de\_Munich}}
  \end{subfigure}
  
  \hspace{-0.5em}
  \begin{subfigure}[b]{0.37\linewidth}
  \centering
    \includegraphics[height=2cm]{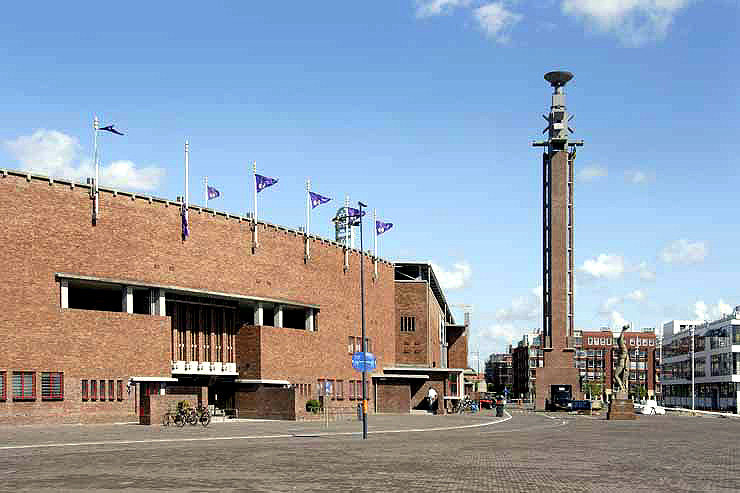}
    \caption{\tiny{\emph{Olympic\_Stadium\_(Amsterdam)}}}
  \end{subfigure}
    \begin{subfigure}[b]{0.3\linewidth}
    \centering
    \includegraphics[height=2cm]{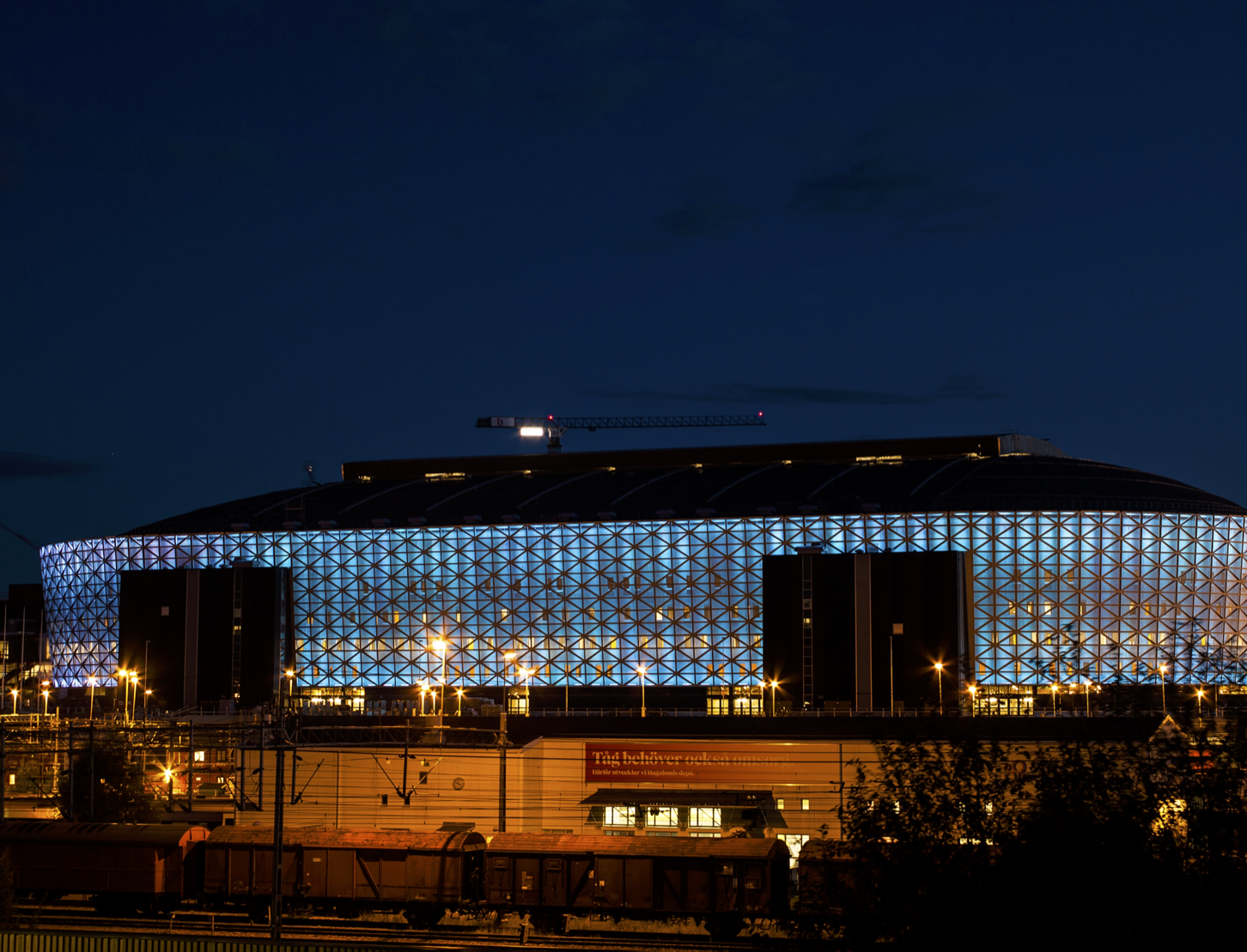}
    \caption{\tiny{\emph{Friends\_Arena}}}
  \end{subfigure}
    \begin{subfigure}[b]{0.32\linewidth}
    \centering
    \includegraphics[height=2cm]{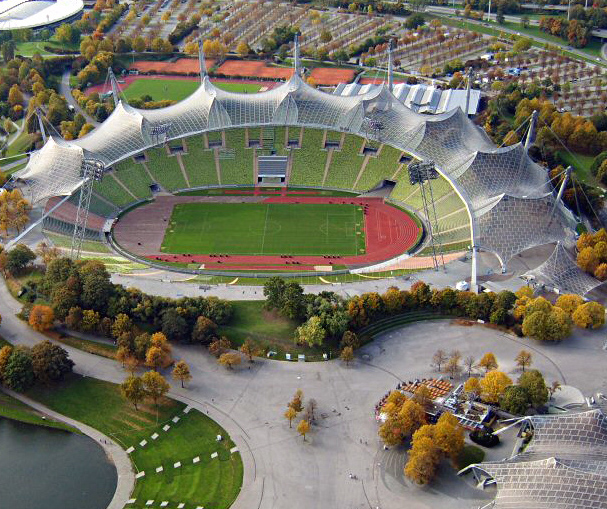}
    \caption{\underline{\tiny{\emph{Olympiastadion\_(Munich)}}}}
  \end{subfigure}
  \caption{\small{\modelname w/o images ranks (b) at top 1, (c) and (d) at top 2 and 3 respectively. Through visual disambiguation, \modelname ranks the correct concept (d) at top 1.}}
  \label{fig:stadium}
\end{figure}

As an example, in DBP15k (FR$\rightarrow$EN), the long-tail entity \emph{Stade\_olympique\_de\_Munich} has only three occurrences in French. 
The top three retrieved entities in English by \modelname w/o visual representation are \emph{Olympic\_Stadium\_(Amsterdam)}, \emph{Friends\_Arena} and \emph{Olympiastadion\_(Munich)}. The embedding without visual information 
was only able to narrow down the answer to European stadiums, but failed to correctly order the specific stadiums (\Cref{fig:stadium}).
With the visual cues, \modelname is able to rank the correct item as the top 1.

Note that in \Cref{fig:long_tail_five_splits}, the split of the most frequent entities (80-100\% quantiles)
generally displays worse performance than the second most frequent split (60-80\% quantiles), suggesting that, 
a denser neighbourhood does not always 
lead to better alignment.
This is consistent with \citet{sun2020alinet}'s observation that, entities with high degree centrality may be affected by the heterogeneity of their neighbourhood.


\subsection{Error Analysis}
\label{sec:error_analysis}

On the monolingual setting (DBP15k-WD), \modelname has reached near-perfect performance. On the cross-lingual setting (DBP15k), however, there is still a $>$20\% gap from perfect alignment. One might wonder why the involvement of images has not solved the remaining 20\% errors. 
By looking into the errors made by \modelname (\textsc{w/o il}) on DBP15k (FR$\rightarrow$EN), we observe that among the 2955 errors, 1945 (i.e., ca. 2/3) of them are entities without valid images. In fact, only 50-70\% of entities in our study have images, according to \Cref{Table:image_coverage_stat}. This is 
inherent to knowledge bases themselves and cannot be easily resolved without an extra step of linking the entities to some external image database. For the remaining 1k errors, ca. 40\% were wrongly predicted regardless of with or without images. 
The other 60\% were correctly predicted before injecting visual information, 
but were missed when images were present. 
Such errors can be mainly attributed to the consistency/robustness issues in visual representations especially for more abstract entities as they tend to have multiple plausible visual representations.
Here is a real example: \emph{Universit\'e\_de\_Varsovie} (French) has the photo of its front gate in the profile while its English equivalent \emph{University\_of\_Warsaw} uses its logo in the profile. The drastically different visual representations cause a misalignment. 
While images are in most cases helpful for the alignment, it requires further investigation for a mechanism to filter out the small fraction of unstable visual representations. This is another substantial 
research direction for future work.

\section{Conclusion}
\label{sec:conclusion}

We propose a new model \modelname that uses images as pivots for aligning entities in different KGs. Through an attention-based modality weighting scheme, we fuse multi-modal information from KGs into a joint embedding and allow the alignment model to automatically adjust modality weights. Besides experimenting with the traditional semi-supervised setting, we present an unsupervised approach, where \modelname leverages visual similarities of entities to build a seed dictionary from scratch and expand the dictionary with iterative learning. The semi-supervised \modelname claims new SOTA on two EA benchmarks, surpassing previous methods by large margins. The unsupervised \modelname achieves $>$70\% accuracy, being close to its performance under the semi-supervised setting, and outperforming the previous best semi-supervised baseline. Finally, we conduct thorough ablation studies and error analysis, offering insights on the benefits of incorporating images for long-tail KG entities. The implication of our work is that perception is a crucial element in learning entity representation and associating knowledge. In this way, our work also highlights the necessity of fusing different modalities in developing intelligent learning systems \citep{mooney2008learning}. 

\section*{Acknowledgement}
We appreciate the anonymous reviewers for their insightful comments and suggestions.
Fangyu Liu is supported by Grace \& Thomas C.H. Chan Cambridge Scholarship. 
This research is supported in part by the Office of the Director of National Intelligence (ODNI), Intelligence Advanced Research Projects Activity (IARPA), via IARPA Contract No. 2019-19051600006 under the BETTER Program, and by Contracts HR0011-18-2-0052 and FA8750-19-2-1004 with the US Defense Advanced Research Projects Agency (DARPA). The views expressed are those of the authors and do not reflect the official policy or position of the Department of Defense or the U.S. Government.

\newpage

\fontsize{9.0pt}{10.0pt}\selectfont

\bibliography{main.bib}
\bibstyle{aaai21}

\clearpage
\appendix


\section{More Implementation Details}\label{sec:implementation}
\Cref{Table:search_space} lists hyper-parameter search space for obtaining the set of used numbers. Instead of always choosing the best performing model, we balance the memory limit and model performance. We train \& evaluate all our models on a machine with the specifications listed in \Cref{Table:hardware}. On this machine, the full training process of \modelname (1,000 epochs) takes 10-15 minutes, depending on the data set. The full model for DBP15k has ca. 16M; for DWY15k has ca. 13M trainable parameters. The difference comes from the different sizes of entity embedding layers. The numbers reported in this paper in general are highly stable and should be easily replicated using our provided code. DBP15k and DWY15k are open benchmark data sets. 

\begin{table}[h!] 
\small
\setlength{\tabcolsep}{1pt}
\caption{Hardware specifications of the used machine.}
\label{Table:hardware}
\centering
\begin{tabular}{lr}
\toprule
hardware & specification \\
\midrule
 RAM & 192 GB \\
 CPU & AMD\textsuperscript{\textregistered} Ryzen 9 3900X 12-core 24-thread \\
 GPU & NVIDIA\textsuperscript{\textregistered} GeForce RTX 2080 Ti (11 GB) $\times$ 2\\
\bottomrule
\end{tabular}
\end{table}

\begin{table}[h!] 
\small
\setlength{\tabcolsep}{1pt}
\caption{This table lists the search space for hyper-parameters used.}
\label{Table:search_space}
\centering
\begin{tabular}{lr}
\toprule
hyper-parameters & search space \\
\midrule
learning rate & \{\texttt{1e-3}, \texttt{5e-4}, \texttt{1e-4}\} \\
\textsc{Gcn} input \& hidden dimension & \{100, 200, 400\}\\
feature dimension of each modality & \{50, 100, 200\} \\
training epochs (before \textsc{il}) & \{300, 500, 1000\} \\
training epochs (total) & \{1000, 1500, 2000\} \\
$K_e$ (for \textsc{il}) & \{3, 5, 10\} \\
$K_s$ (for \textsc{il}) & \{3, 5, 10\} \\
$\alpha$ in \Cref{eq:loss} & \{5, 10, 15, 20\}\\
$\beta$ in \Cref{eq:loss} & \{5, 10, 15, 20\}\\
$k$ in \textsc{Csls} & \{1, 3, 5, 10\}\\
\bottomrule
\end{tabular}
\end{table}

\begin{table*}[h!] 
\small
\setlength{\tabcolsep}{3.2pt}
\caption{Quantitative results on DBP15k. Unsupervised setting.}
\label{Table:unsupervised_dbp15k_full}
\centering
\begin{tabular}{clccccccccccccccccccccccccccccc}
\toprule
 \multirow{2}{*}{seed}  &  \multicolumn{3}{c}{FR$\rightarrow$EN} & $\ $ & \multicolumn{3}{c}{JA$\rightarrow$EN} & $\ $ &  \multicolumn{3}{c}{ZH$\rightarrow$EN} \\
 \cmidrule{2-4}\cmidrule{6-8} \cmidrule{10-12}
   &  {\scriptsize H@1} &\scriptsize H@10 &\scriptsize MRR & &\scriptsize H@1 &\scriptsize H@10 &\scriptsize MRR & &\scriptsize H@1 &\scriptsize H@10 &\scriptsize MRR \\
  \midrule
 100    & .600 \scriptsize $\pm.008$ & .819 \scriptsize $\pm.009$ & .676 \scriptsize $\pm.008$ & & .576 \scriptsize $\pm.005$ & .783 \scriptsize $\pm.010$ & .648 \scriptsize $\pm.006$ & & .609 \scriptsize $\pm.007$ & .809 \scriptsize $\pm.009$ & .679 \scriptsize $\pm.007$ \\
 1,000  & .656 \scriptsize $\pm.005$ & .856 \scriptsize $\pm.002$ & .725  \scriptsize $\pm.003$ & & .624 \scriptsize $\pm.003$ & .817 \scriptsize $\pm.014$ & .690 \scriptsize $\pm.005$ & & .626 \scriptsize $\pm.007$ & .798 \scriptsize $\pm.008$ & .686 \scriptsize $\pm.008$\\ 
 2,000  & .692 \scriptsize $\pm.004$ & .879 \scriptsize $\pm.003$ & .756  \scriptsize $\pm.001$ & & .671 \scriptsize $\pm.011$ & .845 \scriptsize $\pm.020$ & .731 \scriptsize $\pm.013$ & & .657 \scriptsize $\pm.003$ & .820 \scriptsize $\pm.007$ & .714 \scriptsize $\pm.005$\\ 
 3,000  & .720 \scriptsize $\pm.005$ & .897 \scriptsize $\pm.006$ & .781  \scriptsize $\pm.001$ & & .703 \scriptsize $\pm.014$ & .864 \scriptsize $\pm.017$ & .759 \scriptsize $\pm.015$ & & .683 \scriptsize $\pm.006$ & .837 \scriptsize $\pm.008$ & .737 \scriptsize $\pm.006$\\
 4,000  & \textbf{.731} \scriptsize $\pm.004$ & \underline{\textbf{.909}} \scriptsize $\pm.003$ & \textbf{.792}  \scriptsize $\pm.003$ & & .715 \scriptsize $\pm.006$ & .868 \scriptsize $\pm.012$ & .769 \scriptsize $\pm.008$ & & .710 \scriptsize $\pm.005$ & .859 \scriptsize $\pm.006$ & .762 \scriptsize $\pm.004$\\ 
 5,000  & .726 \scriptsize $\pm.006$ & .901 \scriptsize $\pm.003$ & .786  \scriptsize $\pm.007$ & & .727 \scriptsize $\pm.003$ & .881 \scriptsize $\pm.003$ & .782 \scriptsize $\pm.002$ & & .735 \scriptsize $\pm.005$ & .882 \scriptsize $\pm.004$ & .787 \scriptsize $\pm.004$ \\  
 6,000  & \textbf{.731} \scriptsize $\pm.004$ & .903 \scriptsize $\pm.002$ & .791  \scriptsize $\pm.005$ & & \underline{\textbf{.737}} \scriptsize $\pm.008$ & \underline{\textbf{.890}} \scriptsize $\pm.004$ & \underline{\textbf{.791}} \scriptsize $\pm.006$ & & \underline{\textbf{.752}} \scriptsize $\pm.006$ & \underline{\textbf{.895}} \scriptsize $\pm.004$ & \underline{\textbf{.804}} \scriptsize $\pm.005$ \\
 7,000  & .708 \scriptsize $\pm.002$ & .896 \scriptsize $\pm.001$ & .773  \scriptsize $\pm.002$ & & .672 \scriptsize $\pm.020$ & .859 \scriptsize $\pm.010$ & .738 \scriptsize $\pm.017$ & & .704 \scriptsize $\pm.003$ & .870 \scriptsize $\pm.011$ & .763 \scriptsize $\pm.005$\\
\midrule
 \scriptsize supervised & .793 \scriptsize $\pm.005$ & .942 \scriptsize $\pm.002$ & .847 \scriptsize $\pm.004$ & & .762 \scriptsize $\pm.008$ & .913 \scriptsize $\pm.003$ & .817 \scriptsize $\pm.006$ & & .761 \scriptsize $\pm.008$ & .907 \scriptsize $\pm.005$ & .814 \scriptsize $\pm.006$ \\
\bottomrule
\end{tabular}
\end{table*}

 \begin{figure*}[!ht]
     \centering
     \begin{subfigure}{0.48\linewidth}
     \includegraphics[width=\linewidth]{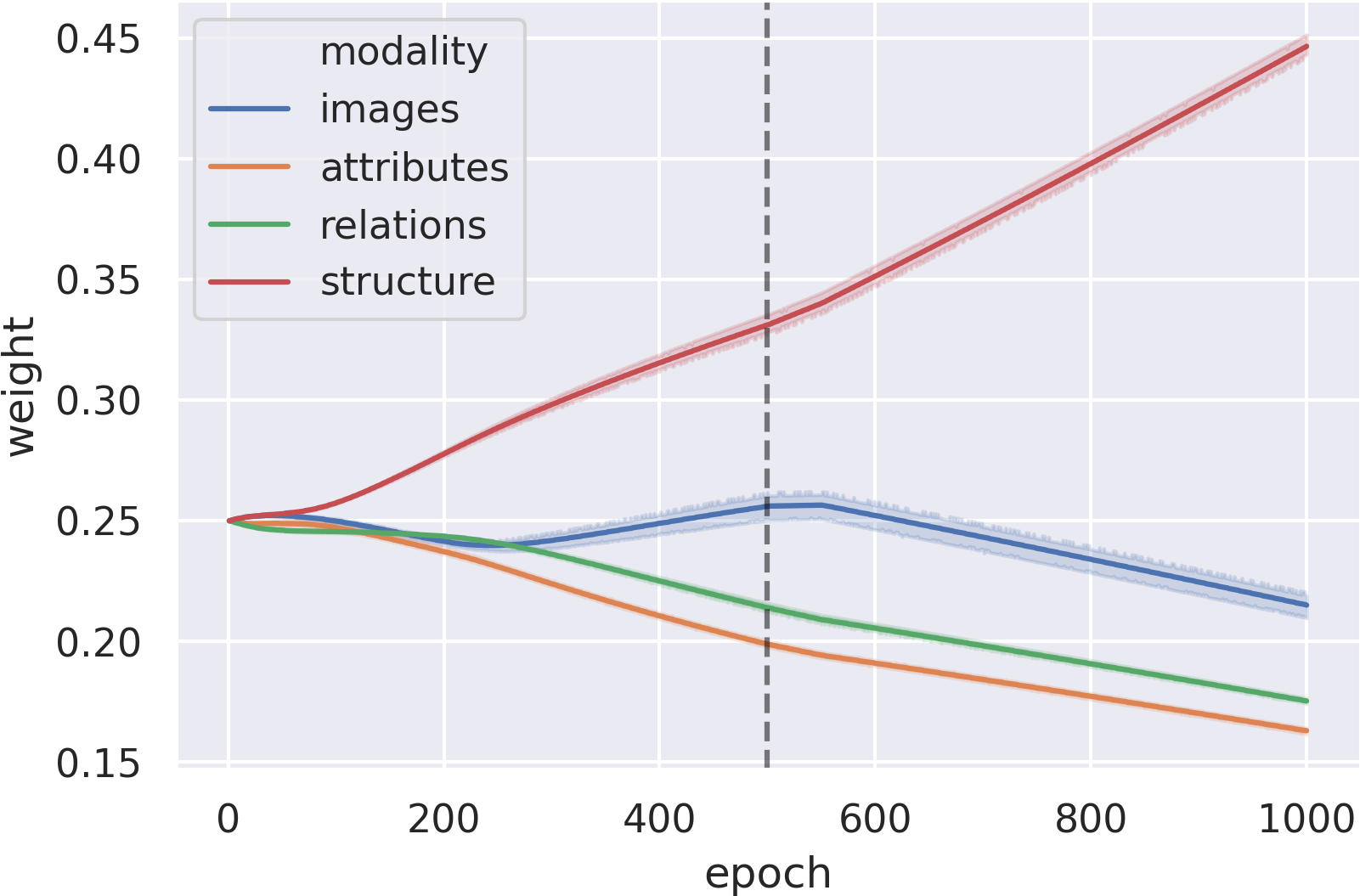}
     \caption{Normalised weights on DBP15k.}
     \end{subfigure}
      \begin{subfigure}{0.48\linewidth}
      \includegraphics[width=\linewidth]{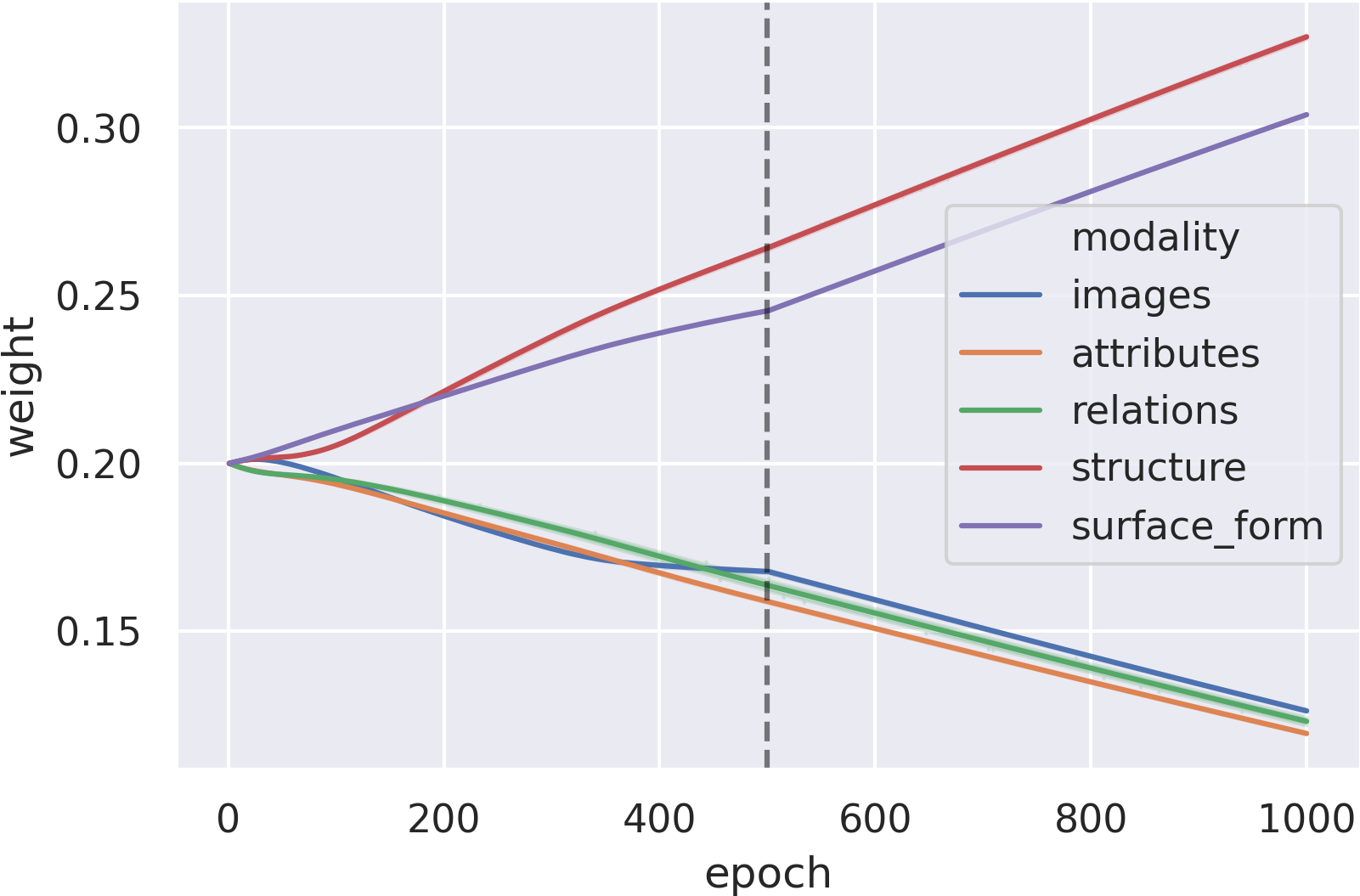}
      \caption{Normalised weights (\textsc{w/ sf}) on DWY15k.}
     \end{subfigure}   
     \caption{Normalised weights against number of epochs.}
     \label{fig:weights}
 \end{figure*}

\section{Full Table for the Unsupervised Setting Results}\label{sup:unsup_table}

\Cref{Table:unsupervised_dbp15k_full} is the same data as \Cref{fig:unsup} but presented using a table.

\section{Plottings of Normalised Modality Weights}\label{sup:modality_weights}

We plot the change of normalised modality weights throughout the training process in \Cref{fig:weights}. It is shown that on DBP15k, images are the second important (after graph structure); on DWY15k (norm), they are the third important (after graph structure and surface form), for almost the whole time of training. Interestingly, on DBP15k, images' weight slightly increases a bit after the starting phase, then starts to decrease once entering iterative learning.

\section{Descriptions of Baseline Methods}\label{sup:baseline}

The baseline methods for cross-lingual EA are in two categories: with or without iterative learning (\textsc{il}). 

Ten of those are without \textsc{il}. Specifically, \textsc{MTransE} \cite{chen2017multigraph} represents a pioneering method of this topic. It jointly learns a translational embedding model \cite{bordes2013translating} and an alignment model that captures the correspondence of counterpart entities via transformations or distances of the embedding representations.
Based on this methodology, \citet{wang2018cross} use \textsc{Gcn} \citep{kipf2017semi} to substitute the translational embedding model to better capture the corresponding entities based on their neighbourhood structures.
\textsc{MuGnn} \citep{cao2019multi} combines multiple channels of GNNs to achieve entity representations that are more robust to parameter initialisation.
\textsc{Mecg} \citep{li2019semi} extends the vanilla \textsc{Gcn} with a regularisation term based on relational translation, aiming at differentiating 
neighbouring entities that participate in different relations.
\textsc{Gcn-Je} \citep{wu2019jointly} extends the same architecture with additional embedding calibration on the relation schemata.
To handle the heterogeneity of neighbourhood entities in different KGs, \textsc{AliNet} employs an attention neighbourhood aggregation, with a gated message passing mechanism to cope with the noises caused by heterogeneous neighbourhood information.
Instead of employing a GNN, \textsc{Rsn} \citep{guo2019learning} uses a residual recurrent network, seeking to capture the long-term dependency of entities on relation paths.
Besides, \textsc{Jape} \citep{sun2017cross} leverages entity attributes to enhance the proximity measure of entities, and \textsc{Hman} \citep{yang2019aligning} incorporates weighted combination of different side information excluding visual modalities.

For the three methods that are trained with \textsc{il}, 
\textsc{BootEA} \citep{sun2018bootstrapping} incorporates the basic bootstrapping approach in \textsc{MTransE}.
\textsc{Mmea} \citep{shi2019modeling} and \textsc{Naea} \citep{zhu2019neighborhood} are \textsc{Gcn} and \textsc{Gat} based, respectively, with the latter using additionally mutual nearest neighbour constraint in proposing new alignment labels.

The monolingual EA setting contains several of those methods that have been reported on the cross-lingual setting at above. The additional baseline method is \textsc{Cotsae} \citep{yang2020cotsae}, which employs an iterative co-training method on the structural and attribute views of entities, similar to the learning process that is employed by \textsc{KDCoE} \citep{chen2018co}.

\section{Future Research Directions}\label{sup:future_work}
\stitle{Investigation of alignment difficulties across different language pairs.}
We observe different patterns of model performance for different language pairs. For example, in \Cref{fig:unsup}, H@1 for FR$\rightarrow$EN plateaus much earlier than JA$\rightarrow$EN and ZH$\rightarrow$EN. 
Understanding these cross-lingual differences requires a more thorough investigation into the distributions of images within each language and how such distributions have influenced cross-lingual mapping. 

\stitle{Extending to low-resource languages.} 
Our study so far has only focused on aligning entities between high-resource languages. However, \modelname can be particularly promising for enhancing knowledge representations in low-resource languages which could benefit the most from knowledge synchronisation with other languages. While the KGs and the alignment information in low-resource languages can be very sparse \cite{sun2020benchmark}, \modelname can leverage crucial side information of entities (ie. images) to facilitate the alignment. 
In the same context, it could also be promising to consider combining multiple sources of high-resource knowledge to jointly learn both the alignment and knowledge transfer to a low-resource KG \cite{chen2020transfer}.


\end{document}